\documentclass[preprint, 12pt]{elsarticle}

\usepackage{lipsum}
\usepackage[utf8]{inputenc}
\usepackage[margin=1in]{geometry}
\usepackage{bold-extra}
\usepackage{xcolor}
\usepackage{adjustbox}
\usepackage{diagbox}
\usepackage{appendix}
\usepackage{lineno}
\usepackage{booktabs}

\usepackage{amssymb}
\usepackage{amsthm}
\usepackage{amsmath}
\usepackage{algorithm}
\usepackage{algpseudocode}
\usepackage{graphicx}
\usepackage{times}
\usepackage[titles]{tocloft}
\usepackage{enumerate}
\usepackage{bm}
\usepackage{sidecap}
\usepackage{epstopdf}
\usepackage{pdfpages}
\usepackage{soul}
\usepackage{multicol}
\usepackage{float}
\usepackage{multibib}
\usepackage{geometry}
\usepackage[small,compact]{titlesec}
\usepackage[normal,bf,labelsep=colon,skip=2pt]{caption}
\usepackage{setspace}
\usepackage{subfig}
\usepackage{tocloft}
\usepackage{url}
\usepackage{floatflt}
\usepackage{array}
\usepackage{pgfgantt}
\usepackage{sidecap}
\usepackage{multirow}
\usepackage{arydshln}
\usepackage[hidelinks]{hyperref}

\numberwithin{equation}{section}
\numberwithin{figure}{section}
\numberwithin{table}{section}

\begin{document}

\begin{frontmatter}

\title{Neural Conditional Gaussian Model: Efficient Data Assimilation with Sequential Uncertainty Quantification for Nonlinear Dynamics}
\author[1]{Chuanqi Chen}
\author[2]{Nan Chen\corref{cor1}}
\cortext[cor1]{Corresponding author}
\emailauthor{chuanqichen@uchicago.edu}{C. Chen}
\emailauthor{chennan@math.wisc.edu}{N. Chen}

\address[1]{Department of the Geophysical Sciences, University of Chicago, Chicago, IL 60637}
\address[2]{Department of Mathematics, University of Wisconsin–Madison, Madison, WI 53706}

\begin{abstract}
Data assimilation (DA) in partially observed nonlinear systems requires repeated estimation of both hidden states and their uncertainty. Conventional nonlinear filtering can be computationally expensive and often relies on an explicit dynamical model. 
In this paper, we develop the Neural Conditional Gaussian Model (Neural CGM), a data-driven dynamical model that maps unobserved physical states into a learned latent space, where the dynamics of the latent states follow a conditional Gaussian structure. This structure allows closed-form DA updates in the latent space, while nonlinear encoder-decoder mappings retain the flexibility needed to represent complex dynamics in physical space. A key feature of Neural CGM is its treatment of uncertainty in DA. The latent posterior mean and covariance are mapped to the physical variables through a Gaussian decoder that combines local moment propagation with learned nonlinear corrections. The resulting physical-space posterior statistics depend on the sequentially updated latent distribution rather than on a separate post-processing estimate. The same statistics enter a Gaussian negative log-likelihood during training, so both the posterior mean and uncertainty affect the learned model. Because Neural CGM learns the dynamical model itself, it can also be used for state forecasting and simulation without explicit knowledge of the governing equations. Numerical experiments on benchmark nonlinear dynamical systems show accurate state estimation and forecasting while substantially reducing online assimilation cost relative to conventional nonlinear DA methods. Neural CGM is therefore suited to repeated state estimation, uncertainty quantification, and prediction from partial observations.
\end{abstract}


\end{frontmatter}

\clearpage

\section{Introduction}
\label{sec:introduction}

Complex dynamical systems arising in scientific and engineering applications are often only partially observed, with representative examples including fluid dynamics, geophysical flows, and aerospace systems~\citep{jost2005dynamical, wiggins2003introduction, majda2006nonlinear, chen2023stochastic, vogiatzoglou2025physics, wang2025development}. Available measurements typically provide information about only a subset of the system variables and are often sparse and contaminated by noise. Consequently, forecasting and other downstream tasks require accurate inference of the unobserved variables together with their associated uncertainty. Data assimilation (DA)~\citep{kalman1960new, majda2012filtering, carrassi2018data, law2015data} addresses this problem by combining observations with a dynamical model to estimate the system state and its uncertainty. A broad range of filtering methods has been developed for this purpose. In particular, ensemble-based methods such as the ensemble Kalman filter~\citep{evensen1994sequential, evensen2003ensemble} have been widely used in large-scale geophysical applications, while particle filters~\citep{gordon1993novel} provide a flexible way to represent non-Gaussian posterior distributions. Extended and unscented Kalman filters~\citep{jazwinski2007stochastic,julier1997new} also remain useful in many nonlinear estimation problems. Despite these successes, important difficulties remain for strongly nonlinear, non-Gaussian, and high-dimensional systems. Linearization can limit the accuracy of the extended Kalman filter, Gaussian or low-order statistical approximations may become inadequate, finite ensembles introduce sampling errors, and particle methods can suffer severe degeneracy as the dimension increases. Reducing these errors often requires larger ensembles or particle populations and can substantially increase the computational cost. In addition, all of these methods rely on a forecast model to propagate state uncertainty between observations. If that model is unavailable or contains substantial model error, the forecast distribution can become inaccurate and degrade the subsequent assimilation. This dependence becomes particularly restrictive when only data from the underlying system are available.


When the governing equations are unavailable, incomplete, or too expensive to integrate repeatedly, machine learning provides an alternative way to construct dynamical models directly from data or to build efficient surrogates of existing physical models~\citep{cheng2023machine}. Data-driven dynamical models have been developed in many forms, including closure models~\citep{sanderse2024scientific, wu2024learning, chen2024physics, dong2025data}, reduced-order models~\citep{majda2018strategies, carlberg2013gnat, noack2011reduced}, neural ODEs~\citep{chen2018neural, chen2025neural}, neural operators~\citep{lu2021learning, li2021fourier, zhang2024modno,lin2023b, wu2023comprehensive, chen2022adenkf}, and generative models~\citep{kingma2013auto,goodfellow2020generative, wu2020enforcing, ho2020denoising, song2020score, bao2023score, bao2025nonlinear}. These methods can provide flexible representations of nonlinear dynamics and substantially reduce the cost of state forecasting. However, an efficient forecast model does not by itself make nonlinear DA inexpensive. When the learned dynamics are strongly nonlinear or effectively black-box, propagating state uncertainty through the model may still require large ensembles, particles, or other computationally intensive approximations. A complementary line of work uses machine learning directly in the assimilation procedure. Recurrent architectures can learn state-estimation mappings from sequential observations, as in RNN-based filters~\citep{chen2023recurrent} and Data Assimilation Network~\citep{boudier2023data}, while KalmanNet~\citep{revach2022kalmannet} learns the Kalman gain from data. Encoder-decoder architectures have also been used for latent-space DA~\citep{zhang2026latent} and the integration of multi-domain observations~\citep{cheng2024multi, behnoudfar2026bridging}, and deep learning can reduce the cost of constructing ensembles for ensemble-based DA~\citep{chattopadhyay2023deep}. These approaches can substantially accelerate state estimation, but many are designed primarily around the assimilation procedure rather than an autonomous dynamical model that can also be used for forecasting and simulation. The challenge is therefore to retain the flexibility of a learned nonlinear dynamical model while making sequential state estimation and uncertainty propagation computationally tractable.

In recent work, the Conditional Gaussian Koopman Network (CGKN)~\citep{chen2025cgkn, chen2025discretecgkn} was developed for learning partially observed dynamical systems while retaining efficient data assimilation. It builds on Conditional Gaussian Nonlinear Systems (CGNS)~\citep{chen2023stochastic, chen2018conditional, chen2022conditional, chen2016filtering, chen2024cgnsde}, whose dynamics may depend nonlinearly on the observed variables while remaining linear in the unobserved variables conditional on the observations. This structure admits closed-form DA formulas through the Conditional Gaussian filter (CG filter), reducing the online cost of DA and avoiding finite-ensemble sampling error in the DA update. CGKN extends this idea by learning nonlinear mappings between the unobserved physical variables and latent variables, allowing nonlinear physical dynamics to be represented while preserving conditionally linear latent dynamics. Here, the latent space refers to a learned representation space in which unobserved physical states are mapped to latent states through a nonlinear encoder. This makes analytical latent-space DA possible even when the unobserved physical variables enter the original dynamics nonlinearly. However, an important issue remains in latent DA: forecasting involves a point-to-point mapping from latent to physical space, whereas the DA decoding step acts on a latent posterior distribution. In CGKN, the same decoder is used for predictive latent states and posterior latent means, while physical-space uncertainty is estimated separately through residual regression based on the current observation. The latent posterior covariance is therefore not used directly to determine the physical-space uncertainty.

The Neural Conditional Gaussian Model (Neural CGM) is built around this distinction between point prediction and posterior inference. Predictive latent states are mapped to the physical variables through a forecast decoder, while DA uses a distribution-aware decoder that takes both the latent posterior mean and covariance as inputs. Because a nonlinear decoder does not preserve Gaussianity, the physical-space posterior is represented by combining local moment propagation with learned corrections to its mean and covariance. In this way, uncertainty accumulated through the observation history is carried directly into the physical-space estimate. These posterior statistics also enter a Gaussian negative log-likelihood during training, so the model is trained on both state estimation and uncertainty quantification rather than on the posterior mean alone. Unlike black-box learned dynamical models, which generally still require an external nonlinear DA procedure, Neural CGM retains a closed-form DA update. Unlike learned DA methods designed primarily for state estimation, it also defines an autonomous dynamical model for forecasting and simulation. Once trained, the DA update does not require ensemble or particle propagation, which is especially useful when inference must be repeated and the underlying physical model is unavailable, imperfect, or expensive to run. We test Neural CGM on the Lorenz 63, Lorenz 96, and Kuramoto-Sivashinsky systems, evaluating forecasting, data assimilation, uncertainty quantification, and online computational cost.

The rest of this paper is organized as follows. Section 2 presents the framework of Neural CGM. Three numerical examples are presented in Section 3 to demonstrate the performance of the proposed framework. Section 4 concludes the paper.

\section{Methodology}
\label{sec:methodology}

Consider the discrete-time partially observed dynamical system (PODS) in the general form:
\begin{equation}
\label{eq:PODS}
\begin{aligned}
    \mathbf{x}_{n+1} &= \mathcal{T}_1(\mathbf{x}_n, \mathbf{y}_n) + \boldsymbol{\epsilon}_n \\
    \mathbf{y}_{n+1} &= \mathcal{T}_2(\mathbf{x}_n, \mathbf{y}_n) + \boldsymbol{\eta}_n
\end{aligned}
\end{equation}
where $[\mathbf{x}; \mathbf{y}]$ is the system state, with $\mathbf{x} \in \mathbb{R}^{d_{\mathbf{x}}}$ denoting the unobserved state variables and $\mathbf{y} \in \mathbb{R}^{d_{\mathbf{y}}}$ denoting the observed state variables. The transition maps $\mathcal{T}_1$ and $\mathcal{T}_2$ are generally nonlinear, and $\boldsymbol{\epsilon}_n$ and $\boldsymbol{\eta}_n$ denote stochastic forcing terms. Here and below, ``unobserved'' refers to the online data-assimilation stage. The offline training procedure described in Section~\ref{ssec:NCGM_Learning} assumes access to paired trajectories of both $\mathbf{x}$ and $\mathbf{y}$. 

For the partially observed dynamical system in \eqref{eq:PODS}, state estimation via data assimilation, also known as filtering, aims to estimate the conditional probability distribution $p(\mathbf{x}_n \mid \mathbf{y}_{0:n})$, while state forecasting concerns the distribution $p(\mathbf{x}_{n+m}, \mathbf{y}_{n+m} \mid \mathbf{x}_n, \mathbf{y}_n)$. In practice, when the unobserved state $\mathbf{x}_n$ is unavailable, a forecast can be initialized from a DA estimate $\mathbf{x}_n^{(\textrm{DA})} \sim p(\mathbf{x}_n \mid \mathbf{y}_{0:n})$, after which the future state is predicted according to
$p(\mathbf{x}_{n+m}, \mathbf{y}_{n+m} \mid \mathbf{x}_n^{(\textrm{DA})}, \mathbf{y}_n)$. This DA-initialized forecasting problem differs from a standard initial-value problem because part of the initial state is not directly observed.

The nonlinearity of the transition maps in \eqref{eq:PODS} can make exact nonlinear filtering intractable and practical DA computationally expensive, especially as the state dimension grows. This difficulty also affects DA-initialized forecasting in applications such as weather and ocean forecasting, wildfire propagation, and robotic navigation. To address this issue, we propose a data-driven approach that learns an expressive structured model, called the Neural Conditional Gaussian Model (Neural CGM), for efficient state estimation and forecasting with uncertainty quantification.

\subsection{Neural Conditional Gaussian Model}\label{ssec:NeuralCGM}
The Neural Conditional Gaussian Model (Neural CGM) reads:
\begin{equation}
\label{eq:NCGM}
\begin{aligned}
    \mathbf{z}_{n+1} &= \mathbf{f}_1(\mathbf{y}_{n}) + \mathbf{g}_1(\mathbf{y}_{n})\mathbf{z}_n + \mathbf{w}_n, \\
    \mathbf{y}_{n+1} &=  \mathbf{f}_2(\mathbf{y}_{n}) + \mathbf{g}_2(\mathbf{y}_{n})\mathbf{z}_n + \mathbf{v}_n,
\end{aligned}
\end{equation}
with $\mathbf{z} = \boldsymbol{\varphi}(\mathbf{x})$ and reconstructed physical state $\hat{\mathbf{x}} = \boldsymbol{\psi}(\mathbf{z})$. Here $\mathbf{z} \in \mathbb{R}^{d_{\mathbf{z}}}$ is the latent embedding of the unobserved physical state $\mathbf{x}$. The maps $\boldsymbol{\varphi}$ and $\boldsymbol{\psi}$ are a nonlinear encoder and forecast decoder, respectively, and $\mathbf{f}_1$, $\mathbf{g}_1$, $\mathbf{f}_2$, and $\mathbf{g}_2$ are nonlinear functions of the observed state. The noise terms satisfy $\mathbf{w}_n \sim \mathcal{N}(\mathbf{0}, \mathbf{Q})$ and $\mathbf{v}_n \sim \mathcal{N}(\mathbf{0}, \mathbf{R})$ and are assumed independent of each other, independent across time, and independent of the current state. The maps $\boldsymbol{\varphi}$, $\boldsymbol{\psi}$, $\mathbf{f}_1$, $\mathbf{g}_1$, $\mathbf{f}_2$, and $\mathbf{g}_2$ are parameterized by neural networks and learned from the paired time-series data $\{(\mathbf{x}_n, \mathbf{y}_n)\}_{n=0}^N$ of the original system in \eqref{eq:PODS}. The noise covariances $\mathbf{Q}$ and $\mathbf{R}$ are subsequently estimated from one-step forecast residuals as described in Section~\ref{ssec:NCGM_Learning}. Thus, Neural CGM learns a dynamical model directly from state trajectories without requiring explicit knowledge of the governing equations.

The conditional Gaussian modeling principle underlying \eqref{eq:NCGM} has a substantial history in nonlinear stochastic modeling. Related conditional Gaussian nonlinear systems have been used for multiscale geophysical flows, stochastic reaction-diffusion systems, and reduced models of climate dynamics, and their analytical conditional statistics have supported efficient data assimilation, uncertainty quantification, Lagrangian inference, and high-dimensional statistical computation~\citep{chen2018conditional, chen2014information, chen2016model, chen2017beating, chen2014predicting, chen2022efficient}. Closely related conditionally linear modeling ideas have also appeared in dynamic stochastic superresolution, stochastic superparameterization, and multiscale data assimilation~\citep{branicki2013dynamic, keating2012new, majda2009mathematical, grooms2014stochastic, majda2014new, majda2014blended, deng2025lemda}. The breadth of these applications suggests that conditional Gaussian structure can be a useful modeling constraint in practice, rather than merely a device for analytical tractability. Neural CGM builds on this idea by learning a nonlinear transformation to a latent space in which analytically tractable DA can be retained without imposing conditional linearity directly on the physical variables.

This latent construction leads to a closed-form DA procedure. In \eqref{eq:NCGM}, conditional on the observed trajectory, the latent state $\mathbf{z}_n$ enters the dynamics linearly. Assuming a Gaussian initial conditional distribution for the latent state, the posterior distribution $p(\mathbf{z}_n \mid \mathbf{y}_{0:n})$ remains Gaussian, and its posterior mean and covariance satisfy the analytical recursions~\citep{chen2016filtering, liptser2013statistics}:
\begin{equation}
\label{eq:CGFilter}
    \begin{aligned}
    \mathbb{E}[\mathbf{z}_n \mid \mathbf{y}_{0:n}] := \boldsymbol{\mu}^{(\mathbf{z})}_{n} &= \mathbf{f}_1 + \mathbf{g}_1\boldsymbol{\mu}^{(\mathbf{z})}_{n-1} - \mathbf{K}_{n} \left(\mathbf{f}_2+\mathbf{g}_2\boldsymbol{\mu}^{(\mathbf{z})}_{n-1} - \mathbf{y}_{n}\right),\\
    \mathrm{cov}(\mathbf{z}_n \mid \mathbf{y}_{0:n}) := \boldsymbol{\Sigma}^{(\mathbf{z})}_{n} &= \mathbf{g}_1\boldsymbol{\Sigma}^{(\mathbf{z})}_{n-1}\mathbf{g}_1^\mathtt{T} + \mathbf{Q} - \mathbf{K}_{n}\mathbf{g}_2\boldsymbol{\Sigma}_{n-1}^{(\mathbf{z})}\mathbf{g}_1^\mathtt{T},
    \end{aligned}
\end{equation}
where
\[
\mathbf{K}_{n}
=
\mathbf{g}_1 \boldsymbol{\Sigma}^{(\mathbf{z})}_{n-1}\mathbf{g}_2^\mathtt{T}
\left(
\mathbf{g}_2\boldsymbol{\Sigma}^{(\mathbf{z})}_{n-1}\mathbf{g}_2^\mathtt{T}
+
\mathbf{R}
\right)^{-1}
\]
is analogous to the Kalman gain. The recursion in~\eqref{eq:CGFilter} is referred to as the conditional Gaussian filter (CG filter). For notational simplicity, the arguments of $\mathbf{f}_1$, $\mathbf{g}_1$, $\mathbf{f}_2$, and $\mathbf{g}_2$ are omitted in \eqref{eq:CGFilter}; all are evaluated at $\mathbf{y}_{n-1}$. For dense matrices, the dominant operations are covariance multiplications and inversion of the innovation covariance; their cost is polynomial in the latent and observed dimensions. Unlike ensemble- and particle-based methods, the online DA step does not introduce an additional ensemble or particle dimension.

The analytical tractability of the CG filter is the main computational advantage of the structure in~\eqref{eq:NCGM}. Conditional linearity is required only with respect to the latent unobserved state $\mathbf{z}$, while general nonlinear dependence on the observed state $\mathbf{y}$ is retained. The nonlinear encoder and decoder further allow the relationship between the physical state $\mathbf{x}$ and latent state $\mathbf{z}$ to remain flexible, even though the latent transition is affine in $\mathbf{z}$ conditional on $\mathbf{y}$. DA is therefore carried out in the latent space, where the posterior statistics are available analytically, and these statistics are subsequently mapped back to the physical state space.

\begin{figure}[htbp]
    \centering
    \includegraphics[width=1.\linewidth]{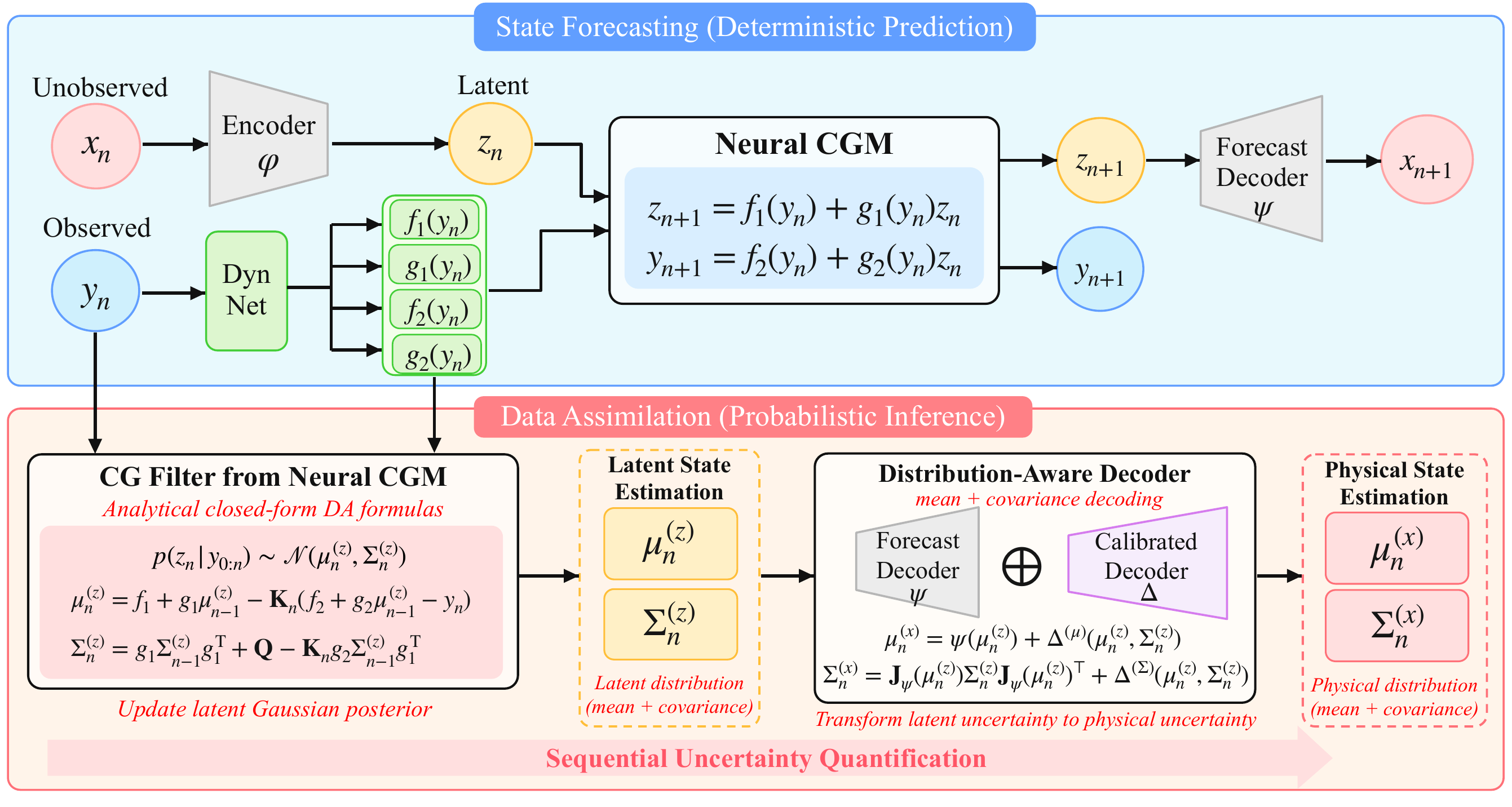}
    \caption{Overview of Neural CGM for point forecasting and distribution-aware data assimilation. The forecasting branch maps predictive latent states to physical states. The DA branch updates the latent posterior mean and covariance analytically and maps both quantities to physical-space posterior statistics, preserving sequential uncertainty information.}
    \label{fig:Schematic}
\end{figure}

\subsection{Neural CGM for Data Assimilation, Forecasting, and Uncertainty Quantification}
\label{ssec:NeuralCGM_Inference}

State forecasting and DA require different latent-to-physical mappings in Neural CGM. In forecasting, a predictive latent state is mapped to a physical state. In DA, however, the latent output is a posterior distribution characterized by both its mean and covariance. The DA decoding step therefore acts on posterior statistics rather than on a single latent state.

After obtaining the latent posterior mean $\boldsymbol{\mu}^{(\mathbf{z})}_n$ and covariance $\boldsymbol{\Sigma}^{(\mathbf{z})}_n$ from the CG filter in~\eqref{eq:CGFilter}, we represent the corresponding physical-space posterior by the Gaussian approximation $q(\mathbf{x}_n\mid \mathbf{y}_{0:n}) = \mathcal{N}\!\left( \boldsymbol{\mu}^{(\mathbf{x})}_n, \boldsymbol{\Sigma}^{(\mathbf{x})}_n \right)$, where the mean and covariance are obtained through a distribution-aware decoder:
\begin{equation}
\label{eq:Gaussian_decoder}
\begin{aligned}
\boldsymbol{\mu}^{(\mathbf{x})}_n
&=
\boldsymbol{\psi}(\boldsymbol{\mu}_n^{(\mathbf{z})})
+
\boldsymbol{\Delta}^{(\boldsymbol{\mu})}
(\boldsymbol{\mu}_n^{(\mathbf{z})},\boldsymbol{\Sigma}_n^{(\mathbf{z})}),
\\
\boldsymbol{\Sigma}^{(\mathbf{x})}_n
&=
\mathbf{J}_{\boldsymbol{\psi}}(\boldsymbol{\mu}_n^{(\mathbf{z})})
\boldsymbol{\Sigma}_n^{(\mathbf{z})}
\mathbf{J}_{\boldsymbol{\psi}}(\boldsymbol{\mu}_n^{(\mathbf{z})})^{\top}
+
\boldsymbol{\Delta}^{(\boldsymbol{\Sigma})}
(\boldsymbol{\mu}_n^{(\mathbf{z})},\boldsymbol{\Sigma}_n^{(\mathbf{z})}).
\end{aligned}
\end{equation}
Here $\mathbf{J}_{\boldsymbol{\psi}}(\boldsymbol{\mu}_n^{(\mathbf{z})})$ denotes the Jacobian matrix of the decoder $\boldsymbol{\psi}$ evaluated at $\boldsymbol{\mu}_n^{(\mathbf{z})}$:
\begin{equation*}
\mathbf{J}_{\boldsymbol{\psi}}(\boldsymbol{\mu}_n^{(\mathbf{z})})
:=
\left.
\frac{\partial \boldsymbol{\psi}(\mathbf{z})}
{\partial \mathbf{z}}
\right|_{\mathbf{z}=\boldsymbol{\mu}_n^{(\mathbf{z})}}.
\end{equation*}

To ensure that the physical-space covariance $\boldsymbol{\Sigma}^{(\mathbf{x})}$ is positive semidefinite (PSD), the covariance correction can be parameterized as \(\boldsymbol{\Delta}^{(\boldsymbol{\Sigma})} = \mathbf{L}\mathbf{L}^\top\). Since both \(\mathbf{J}_{\psi}\boldsymbol{\Sigma}^{(\mathbf{z})}\mathbf{J}_{\psi}^{\top}\) and \(\mathbf{L}\mathbf{L}^{\top}\) are PSD, their sum is also PSD. To further guarantee positive definiteness (PD), which is required for the covariance inverse and log-determinant in the Gaussian likelihood to be well defined during the learning process (see Section~\ref{ssec:NCGM_Learning}), a small diagonal regularization term \(\epsilon\mathbf{I}\) can be added, where \(\epsilon>0\). The physical-space covariance is therefore parameterized as $ \boldsymbol{\Sigma}^{(\mathbf{x})} = \mathbf{J}_{\psi}\boldsymbol{\Sigma}^{(\mathbf{z})}\mathbf{J}_{\psi}^{\top} + \mathbf{L}\mathbf{L}^{\top} + \epsilon\mathbf{I}$,  which guarantees that \(\boldsymbol{\Sigma}^{(\mathbf{x})}\) is PD.

The distinction between point-state decoding and posterior decoding is important when $\boldsymbol{\psi}$ is nonlinear. The exact pushforward of a Gaussian latent posterior through a nonlinear decoder is generally non-Gaussian, and $\mathbb{E}[\boldsymbol{\psi}(\mathbf{z})]\neq \boldsymbol{\psi}(\mathbb{E}[\mathbf{z}])$ in general. For example, a second-order Taylor expansion gives, componentwise,
\begin{equation*}
\mathbb{E}[\psi_j(\mathbf{z})]
\approx
\psi_j(\boldsymbol{\mu}^{(\mathbf{z})})
+
\frac{1}{2}
\operatorname{tr}\!\left(
\mathbf{H}_{\psi_j}(\boldsymbol{\mu}^{(\mathbf{z})})
\boldsymbol{\Sigma}^{(\mathbf{z})}
\right),
\end{equation*}
where $\mathbf{H}_{\psi_j}$ is the Hessian of the $j$-th decoder component. This illustrates why the physical-space posterior mean can depend on both the latent mean and covariance. Similarly, the Jacobian term in~\eqref{eq:Gaussian_decoder} gives the first-order covariance transformation, while $\boldsymbol{\Delta}^{(\boldsymbol{\mu})}$ and $\boldsymbol{\Delta}^{(\boldsymbol{\Sigma})}$ learn corrections for nonlinear effects not represented by the local approximation.

An important consequence is that uncertainty in physical space remains connected to the sequential DA process. Both $\boldsymbol{\mu}^{(\mathbf{z})}_n$ and $\boldsymbol{\Sigma}^{(\mathbf{z})}_n$ depend on the observation history $\mathbf{y}_{0:n}$ through the CG-filter recursion, and both enter the physical-space decoder. Thus, the uncertainty estimate is obtained from the evolving latent posterior rather than being introduced only after state estimation.

For state forecasting, given a complete initial state $[\mathbf{x}_0; \mathbf{y}_0]$, future predictions $[\tilde{\mathbf{x}}_n; \tilde{\mathbf{y}}_n]$ are obtained by recursively evaluating the Neural CGM model in~\eqref{eq:NCGM}. When the initial unobserved state $\mathbf{x}_0$ is unavailable and only an observation history $\mathbf{y}_{-m:0}$ is given, DA can first be performed using the CG filter in~\eqref{eq:CGFilter} and the distribution-aware decoder in~\eqref{eq:Gaussian_decoder} to construct $q(\mathbf{x}_0\mid\mathbf{y}_{-m:0})$. Its posterior mean, or samples drawn from this approximate posterior, can then be used to initialize future forecasts. Neural CGM therefore provides an autonomous learned dynamical model together with DA and uncertainty quantification, allowing the same model to be used for state estimation, forecasting, and simulation when part of the system is unobserved.

\subsection{Learning Neural CGM from Data}
\label{ssec:NCGM_Learning}

Neural CGM is trained offline using paired time-series data
$\{(\mathbf{x}_n,\mathbf{y}_n)\}_{n=0}^N$
from the system in~\eqref{eq:PODS}. Here $\mathbf{x}$ is unobserved during the online DA stage but is assumed to be available in the offline training data. In other words, full-state trajectories, including both observed and unobserved variables, are required only during offline training, whereas online data assimilation uses only observed variables. Such training data may come from reanalysis datasets, numerical simulations or laboratory measurements. The encoder $\boldsymbol{\varphi}$, forecast decoder $\boldsymbol{\psi}$, nonlinear functions $\mathbf{f}_1$, $\mathbf{g}_1$, $\mathbf{f}_2$, and $\mathbf{g}_2$, and the correction networks $\boldsymbol{\Delta}^{(\boldsymbol{\mu})}$ and $\boldsymbol{\Delta}^{(\boldsymbol{\Sigma})}$ are learned by minimizing
\begin{equation}
\label{eq:risk}
\mathcal{R}
=
\lambda_{\mathrm{AE}}\mathcal{R}_{\mathrm{AE}}
+
\lambda_{\mathrm{forecast}}\mathcal{R}_{\mathrm{forecast}}
+
\lambda_{\mathrm{DA}}\mathcal{R}_{\mathrm{DA}}.
\end{equation}
The three terms target the latent representation, dynamical prediction, and DA posterior, respectively:
\begin{align}
\mathcal{R}_{\textrm{AE}}
&:=
\mathbb{E}_{\mathbf{x}}
\left[
\|\mathbf{x}
-
\boldsymbol{\psi}(\boldsymbol{\varphi}(\mathbf{x}))\|^2
\right],
\label{eq:Risk_AE}\\
\mathcal{R}_{\textrm{forecast}}
&:=
\mathbb{E}_{(\mathbf{x}_0,\mathbf{y}_0)}
\left[
\frac{1}{N_s}
\sum_{n=1}^{N_s}
\left\{
\|\mathbf{x}_n-\tilde{\mathbf{x}}_n\|^2
+
\|\mathbf{y}_n-\tilde{\mathbf{y}}_n\|^2
+
\|\mathbf{z}_n-\tilde{\mathbf{z}}_n\|^2
\right\}
\right],
\label{eq:Risk_forecast}\\
\mathcal{R}_{\textrm{DA}}
&:=
\mathbb{E}_{(\mathbf{x}_{0:N_l},\mathbf{y}_{0:N_l})}
\left[
\frac{1}{N_l-N_c}
\sum_{n=N_c+1}^{N_l}
\left\{
\left(\mathbf{x}_n-\boldsymbol{\mu}_n^{(\mathbf{x})}\right)^\top
\left(\boldsymbol{\Sigma}_n^{(\mathbf{x})}\right)^{-1}
\left(\mathbf{x}_n-\boldsymbol{\mu}_n^{(\mathbf{x})}\right)
+
\log\det\!\left(\boldsymbol{\Sigma}_n^{(\mathbf{x})}\right)
\right\}
\right].
\label{eq:Risk_DA}
\end{align}
Here $\lambda_{\textrm{AE}}$, $\lambda_{\textrm{forecast}}$, and $\lambda_{\textrm{DA}}$ are the weights of the three loss terms, and $\|\cdot\|$ denotes the standard vector $\ell^2$-norm. The hyperparameters $N_s$, $N_l$, and $N_c$ denote the forecast horizon, DA window length, and DA warm-up length, respectively. The variables $\tilde{\mathbf{x}}_n$, $\tilde{\mathbf{y}}_n$, and $\tilde{\mathbf{z}}_n$ are the forecast states, while $\boldsymbol{\mu}_n^{(\mathbf{x})}$ and $\boldsymbol{\Sigma}_n^{(\mathbf{x})}$ are the physical-space posterior statistics defined in~\eqref{eq:Gaussian_decoder}. 

The DA loss in~\eqref{eq:Risk_DA} is proportional, up to an additive constant and an overall factor, to the negative Gaussian log-likelihood of the approximate physical-space posterior. The quadratic term scales the state-estimation error by the predicted covariance, while the log-determinant term prevents the likelihood from being improved simply by inflating that covariance. The loss therefore trains the posterior mean and uncertainty together rather than optimizing the posterior mean alone. In the present implementation, only the diagonal entries of $\boldsymbol{\Sigma}_n^{(\mathbf{x})}$ are retained in this likelihood, which reduces the computational cost and focuses the uncertainty quantification on the marginal uncertainty of each state variable. If computational resources permit, a full covariance representation could be adopted when cross-variable correlations are of interest or the diagonal approximation proves insufficient.

After the neural components are trained, the covariance matrices $\mathbf{Q}$ and $\mathbf{R}$ of the stochastic terms in~\eqref{eq:NCGM} are estimated from one-step forecast residuals:
\begin{equation}
\label{eq:cov_est}
\begin{aligned}
\mathbf{Q}
&=
\frac{1}{N}
\sum_{n=1}^{N}
\bigl(\mathbf{z}_{n}-\tilde{\mathbf{z}}_{n}\bigr)
\bigl(\mathbf{z}_{n}-\tilde{\mathbf{z}}_{n}\bigr)^\top,
\\
\mathbf{R}
&=
\frac{1}{N}
\sum_{n=1}^{N}
\bigl(\mathbf{y}_{n}-\tilde{\mathbf{y}}_{n}\bigr)
\bigl(\mathbf{y}_{n}-\tilde{\mathbf{y}}_{n}\bigr)^\top.
\end{aligned}
\end{equation}
Here $\tilde{\mathbf{z}}_{n}$ and $\tilde{\mathbf{y}}_{n}$ denote the corresponding one-step predictions from $\mathbf{z}_{n-1}$ and $\mathbf{y}_{n-1}$. In this work, $\mathbf{Q}$ and $\mathbf{R}$ are taken to be diagonal, and only their variance terms are estimated.

The autoencoder loss constructs the latent representation, the forecasting loss trains the dynamical model and its forecast decoder, and the DA likelihood trains the physical-space posterior produced by sequential assimilation. The learned model is therefore shaped not only by trajectory prediction but also by state estimation and uncertainty quantification.

Figure~\ref{fig:Schematic} highlights the distinction between point-state forecasting and posterior inference in Neural CGM. This distinction is also what separates the proposed formulation from CGKN \cite{chen2025cgkn}. In CGKN, a shared decoder is used for predictive latent states and DA posterior means, while physical-space uncertainty is estimated separately through residual regression. Neural CGM instead carries both the latent posterior mean and covariance into a distribution-aware physical-space decoder and trains the resulting posterior statistics through a Gaussian likelihood. Consequently, the physical-space uncertainty remains connected to the sequential DA update and to the full observation history. At the same time, the forecast decoder is free to focus on point prediction, so the learned model retains its role as an autonomous dynamical model for forecasting and simulation. These features allow Neural CGM to combine closed-form latent-space DA, sequential uncertainty quantification, and nonlinear state forecasting within the same learned dynamical model.

\section{Numerical Experiments}
\label{sec:NumericalExperiment}

\begin{table}[htbp] 
    \centering
    \caption{Benchmark of different methods for state forecasting and data assimilation on various systems.}
    \label{tab:benchmarks}
    \renewcommand{\arraystretch}{1.2}
    \begin{adjustbox}{max width=1.\textwidth,center}
    \begin{tabular}{c|c|rrrrr}
        \toprule
        System & Method & Forecast RMSE & DA RMSE & DA NLL & DA Time (s) & Parameters \\
        \hline
        \hline

        \multirow{7}{*}{
            \begin{tabular}{c}
                \textbf{Lorenz 63} \\
                \textbf{system} \\
                \\
                $\begin{aligned}
                \text{State dim} &= 3 \\
                \text{{Obs dim}} &= 1 \\
                \text{Unobs dim} &= 2
                \end{aligned}$ \\
            \end{tabular}
        }
          & Neural CGM & 1.6907 & 1.3882 & 3.8068
          & 0.8772 & 23,387 \\
          & CGKN & 1.6941 & 1.4174 & 3.9126
          & 0.9367 & 25,737 \\
          & DNN & 1.6887 & --- & --- & --- & 25,219 \\
          & EnKF & --- & 1.4594 & 3.1563
          & 16.0968 & Governing Eq. \\
          & PF & --- & 1.3027 & 2.8322
          & 26.3887 & Governing Eq. \\
          & EKF & --- & 5.2498 & 48.7082
          & 46.5615 & Governing Eq. \\
          & UKF & --- & 1.4925 & 3.6059
          & 38.3962 & Governing Eq. \\
        \hline

        \multirow{7}{*}{
            \begin{tabular}{c}
                \textbf{Lorenz 96} \\
                \textbf{system} \\
                \\
                $\begin{aligned}
                \text{State dim} &= 40 \\
                \text{{Obs dim}} &= 4 \\
                \text{Unobs dim} &= 36
                \end{aligned}$ \\
            \end{tabular}
        }
          & Neural CGM & 0.7353 & 3.1927 & 92.1253
          & 3.5831 & 207,612 \\
          & CGKN & 0.7356 & 3.1962 & 95.2390
          & 2.8629 & 208,884 \\
          & DNN & 0.7394 & --- & --- & --- & 224,680 \\
          & EnKF & --- & 3.2567 & 92.7474
          & 107.8108 & Governing Eq. \\
          & PF & --- & 3.4697 & 95.6056
          & 515.3742 & Governing Eq. \\
          & EKF & --- & $\infty$ & $\infty$
          & - & Governing Eq. \\
          & UKF & --- & 5.1786 & 440.7245
          & 333.9104 & Governing Eq. \\
        \hline

        \multirow{9}{*}{
            \begin{tabular}{c}
                \textbf{Kuramoto-} \\
                \textbf{Sivashinsky} \\
                \textbf{equation} \\
                \\
                $\begin{aligned}
                \text{State dim} &= 128 \\
                \text{{Obs dim}} &= 2 \\
                \text{Unobs dim} &= 126
                \end{aligned}$ \\
            \end{tabular}
        }
          & Neural CGM & 0.0554 & 0.3846 & 54.6653
          & 1.8307 & 108,745 \\
          & CGKN & 0.0621 & 0.4222 & 166.1100
          & 1.3559 & 116,811 \\
          & CNN & 0.0967 & --- & --- & --- & 123,969 \\
          & FNO & 0.0448 & --- & --- & --- & 113,249 \\
          & EnKF & --- & 1.0498 & 189.7641
          & 185.0903 & Governing Eq. \\
          & PF & --- & 0.4347 & 52.4199
          & 706.2211 & Governing Eq. \\
          & EKF & --- & 1.2971 & 289.6295
          & 257.2876 & Governing Eq. \\
          & UKF & --- & 1.2501 & 217.9368
          & 115.1386 & Governing Eq. \\
          & Interpolation & --- & 1.5570  & ---
          & 0.0004 & --- \\
        \hline
        \bottomrule
    \end{tabular}
    \end{adjustbox}
\end{table}

In this section, we evaluate the performance of Neural CGM in terms of state forecasting, data assimilation, uncertainty quantification, and computational efficiency against other methods. The main evaluation metrics are the root mean square error (RMSE) and the negative log-likelihood (NLL). RMSE measures the accuracy of the predicted or estimated states, whereas NLL evaluates the quality of the predicted posterior distribution by accounting for both its mean and covariance.

The DA negative log-likelihood (NLL) is evaluated using the Gaussian posterior predicted for the unobserved physical state:
\[
\mathrm{DA\ NLL}
:=
\frac{1}{2N}
\sum_{n=1}^{N}
\left[
(\mathbf{x}_n-\boldsymbol{\mu}^{(\mathbf{x})}_n)^\top
(\boldsymbol{\Sigma}^{(\mathbf{x})}_n)^{-1}
(\mathbf{x}_n-\boldsymbol{\mu}^{(\mathbf{x})}_n)
+
\log\det\boldsymbol{\Sigma}^{(\mathbf{x})}_n
+
d_{\mathbf{x}}\log(2\pi)
\right],
\]where \(d_\mathbf{x}\) is the dimension of the unobserved physical state and \(N\) is the number of evaluation time steps. For the diagonal covariance used in the present implementation, this expression is evaluated componentwise and summed over the state variables. In this work, only the diagonal entries of \(\boldsymbol{\Sigma}^{(\mathbf{x})}_n\) are used to evaluate the DA NLL reported in Table~\ref{tab:benchmarks}.

All quantitative metrics in Table~\ref{tab:benchmarks} are evaluated over long held-out trajectories rather than a small collection of test samples. The DA statistics use 8000, 20000, and 10000 test time steps for the Lorenz 63, Lorenz 96, and Kuramoto-Sivashinsky examples, respectively. These long trajectories make the time-averaged RMSE and NLL less sensitive to individual episodes, although the samples are temporally correlated. The reported values do not quantify variability from neural-network initialization or training, so very small differences between data-driven models should not be over-interpreted.
\begin{itemize}
\item \textbf{Neural Conditional Gaussian Model (Neural CGM):} This is the method proposed in this work, designed to achieve efficient and accurate state forecasting and data assimilation while providing uncertainty quantification. The training hyper-parameter settings are summarized in Table~\ref{tab:training_hyperparameters}.
\item \textbf{Conditional Gaussian Koopman Network (CGKN):} CGKN \cite{chen2025cgkn} is a previous machine-learning-based framework developed for modeling dynamical systems and enabling efficient data assimilation. Compared with the proposed Neural CGM, CGKN uses a shared decoder to map both predictive latent states from state forecasting and posterior latent means from data assimilation back to the physical state space, rather than explicitly separating these two latent-to-physical mappings. This shared decoding can limit the flexibility of the model when the two tasks require different transformations. Moreover, CGKN does not propagate a physical-space posterior covariance sequentially; instead, uncertainty is estimated through a post-hoc residual analysis using a regression-based approach.
\item \textbf{Classical nonlinear data assimilation methods:} These methods include the extended Kalman filter (EKF), unscented Kalman filter (UKF), ensemble Kalman filter (EnKF), and particle filter (PF). The PF is a bootstrap particle filter with adaptive resampling based on the effective sample size (ESS). In each numerical experiment, the governing equations are assumed to be known. The key parameters and configurations of these classical methods, including sigma-point parameters, covariance inflation, localization settings, resampling, and roughening settings, are specified below for each experiment.
\item \textbf{Pure neural network models:} Deep Neural Network (DNN), Convolutional Neural Network (CNN), and Fourier Neural Operator (FNO) are used as benchmarks for evaluating the expressive capability of the proposed Neural CGM in learning the underlying system dynamics and performing state forecasting. Compared with these neural network models, which are used here primarily for state forecasting, a key characteristic of Neural CGM is its ability to perform efficient data assimilation with uncertainty quantification for state estimation.
\end{itemize}

\begin{table}[htbp]
\centering
\caption{Training hyperparameters of Neural CGM for the three numerical examples reported in Table~\ref{tab:benchmarks}. Here, $d_{\mathbf{z}}$ denotes the latent-state dimension, $N_s$ the state-forecast horizon,
$N_l$ the data assimilation (DA) horizon, and $N_c$ the DA warm-up length excluded from the DA loss evaluation.}
\label{tab:training_hyperparameters}
\begin{tabular}{l*{4}{>{\centering\arraybackslash}p{1.1cm}}}
\hline
System & \multicolumn{4}{c}{Hyperparameters in Neural CGM} \\
\cline{2-5}
& $d_{\mathbf{z}}$ & $N_s$ & $N_l$ & $N_c$ \\
\hline
Lorenz 63 System               & 12 & 1 & 200 & 0  \\
Lorenz 96 System               & 80 & 1 & 100 & 20 \\
Kuramoto--Sivashinsky Equation & 15 & 1 & 200 & 20 \\
\hline
\end{tabular}
\end{table}

All inference computations in each experiment were performed on a MacBook Air with an Apple M2 chip and 16 GB of unified memory. For machine-learning-based model training, the models in Lorenz 63 examples was trained on the same MacBook, while those in examples of Lorenz 96 and Kuramoto-Sivashinsky were trained on an NVIDIA GeForce RTX 5090 GPU with 32 GB of GPU memory.

\subsection{Stochastic Lorenz 63 System}
\label{ssec:L63}

The Lorenz 63 system is a classical three-dimensional chaotic dynamical system and is widely used as a benchmark for studies of chaos and nonlinear data assimilation. In this example, we consider a stochastic Lorenz 63 system with additive dynamical noise:
\begin{equation}
\label{eq:L63}
\begin{aligned}
    \mathrm{d}x &= \sigma(y-x)\,\mathrm{d}t + \sigma_{x}\,\mathrm{d}W_x, \\
    \mathrm{d}y &= \left[x(\rho-z)-y\right]\,\mathrm{d}t + \sigma_{y}\,\mathrm{d}W_y, \\
    \mathrm{d}z &= (xy-\beta z)\,\mathrm{d}t + \sigma_{z}\,\mathrm{d}W_z,
\end{aligned}
\end{equation}
where $(x,y,z)$ are the system states and $W_x,W_y,W_z$ are independent standard Wiener processes. The parameters are set as $\sigma = 10, \rho = 28, \beta = \frac{8}{3}, \sigma_{x}  = \sigma_{y} = \sigma_{z} = 0.5$. The system is simulated over $L_t = 2000$ time units with numerical time step $\Delta t = 10^{-3}$ using the Euler-Maruyama method with initial condition $(0,0,0)$. The simulated trajectory is sub-sampled every 50 time steps, yielding observational data with temporal resolution $\Delta t_{\mathrm{obs}} = 0.05$. Independent measurement noise following $\mathcal{N}(0,2^2)$ is added to all state variables in the data used by the data-driven models. The first $80\%$ of the trajectory is used as training data and the remaining $20\%$ is used as test data. During the DA test, only the noisy $x$ component is supplied as an observation, while $y$ and $z$ are withheld and treated as unobserved states; the corresponding clean simulated states are used as reference truth for evaluation.

For data assimilation, the proposed Neural CGM is compared with CGKN and several classical DA methods, including the extended Kalman filter (EKF), unscented Kalman filter (UKF), ensemble Kalman filter (EnKF), and particle filter (PF). As data-driven approaches, both Neural CGM and CGKN learn the dynamical model directly from data with comparable model complexity and numbers of parameters. For the classical DA methods, the true governing equations are assumed to be known, and the Lorenz 63 dynamics are propagated between consecutive observations using an internal time step of $10^{-3}$ to ensure numerical stability and accuracy. For the EKF and UKF, the initial mean and covariance are estimated from the training data, whereas the initial ensemble members and particles for the EnKF and PF are randomly sampled from the training data to provide representative prior states. In the UKF, the scaling parameters are set to $\alpha=0.3$, $\kappa=0$, and $\beta=2$, where $\alpha$ and $\kappa$ control the spread of the sigma points, while $\beta=2$ is a standard choice for approximately Gaussian distributions. The EnKF employs $200$ ensemble members, whereas the PF uses $1000$ particles, with resampling triggered when the effective sample size falls below $50\%$ of the particle population.

\begin{figure}[htbp]
    \centering
    \includegraphics[width=\linewidth]{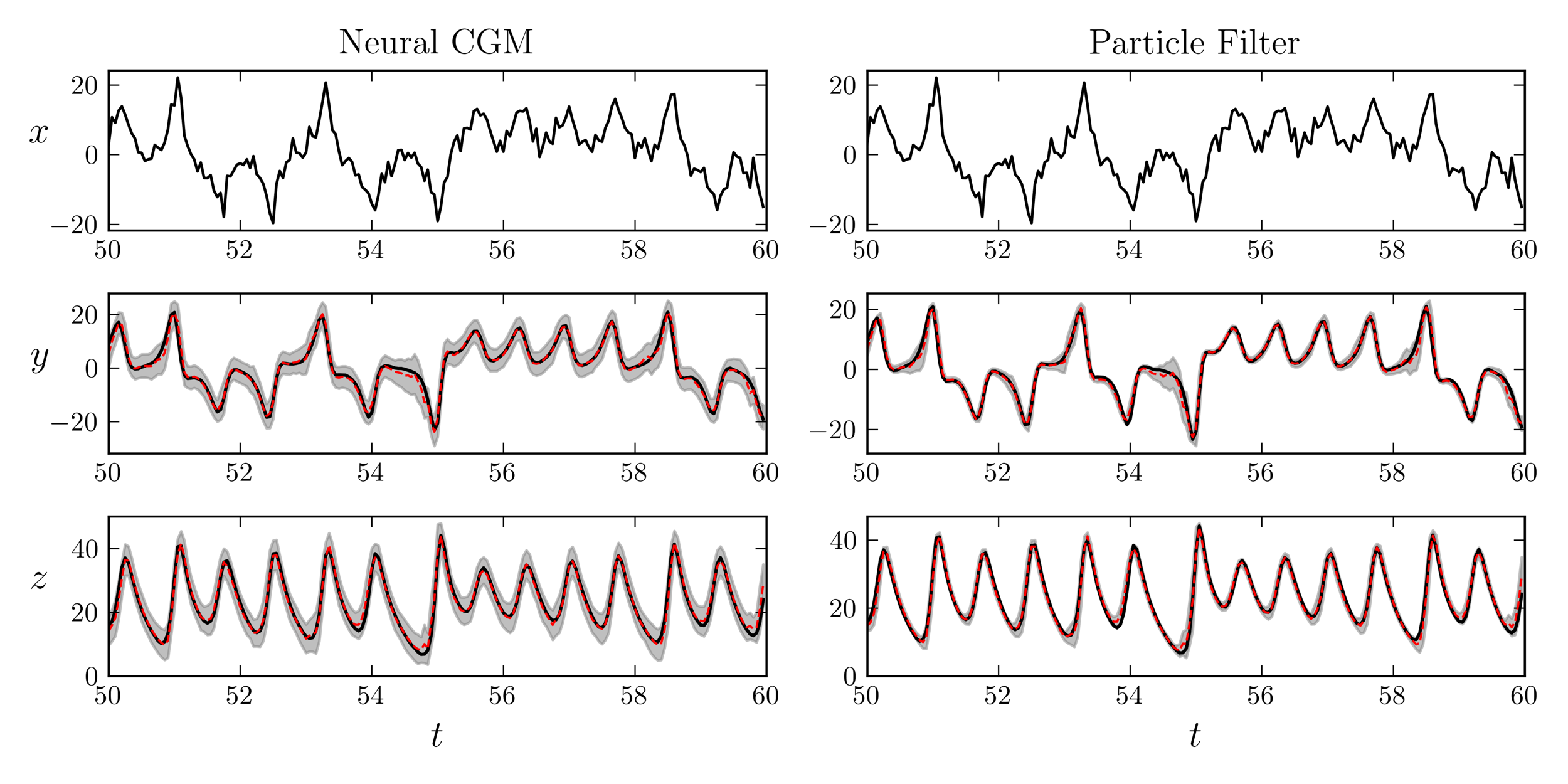}
    \caption{Time series of the data assimilation results for the stochastic Lorenz-63 system. The black trajectory for $x$ represents the noisy observations. For the hidden states $y$ and $z$, the black trajectories denote the true states, while the red trajectories represent the posterior means estimated by the Neural CGM and the particle filter (PF). The gray shaded regions indicate uncertainty intervals corresponding to $\pm 2$ posterior standard deviations.}
    \label{fig:L63_DA_TimeSeries}
\end{figure}

Figure~\ref{fig:L63_DA_TimeSeries} presents the data assimilation results obtained using the Neural CGM and the particle filter (PF). The state variable $x$ is directly observed, while $y$ and $z$ are treated as hidden states. For the observed variable $x$, the black trajectory represents the observed data, i.e., the true state contaminated by measurement noise. For the hidden variables $y$ and $z$, the black trajectories represent the true states, the red trajectories denote the posterior means, and the gray shaded regions indicate the uncertainty intervals corresponding to two posterior standard deviations. Both methods accurately recover the true hidden-state trajectories, with the uncertainty intervals generally covering the true states. The Neural CGM exhibits slightly wider uncertainty intervals than the PF. In terms of computational efficiency, however, the Neural CGM requires only $0.8772$ s to assimilate $8000$ time steps, compared with approximately $26.3887$ s for the PF, corresponding to an approximately $30\times$ speedup.

In this low-dimensional case, Monte-Carlo-based DA methods, such as the particle filter (PF) and ensemble Kalman filter (EnKF), achieve competitive performance and slightly outperform the ML-based methods on some evaluation metrics in Table~\ref{tab:benchmarks}. This is consistent with our expectations, since a sufficiently large number of samples or ensemble members can provide an accurate approximation of the underlying probability distribution in low-dimensional systems. In addition, the ML-based methods are trained on noisy data, which can adversely affect their performance. Nevertheless, the results demonstrate that Neural CGM remains effective when trained on noisy data. In the following two higher-dimensional nonlinear systems, the computational advantages of the proposed method become more evident. By exploiting an analytical filtering update in latent space, Neural CGM achieves competitive or improved estimation accuracy while substantially reducing the online assimilation cost.

\begin{figure}[htbp]
    \centering
    \includegraphics[width=\linewidth]{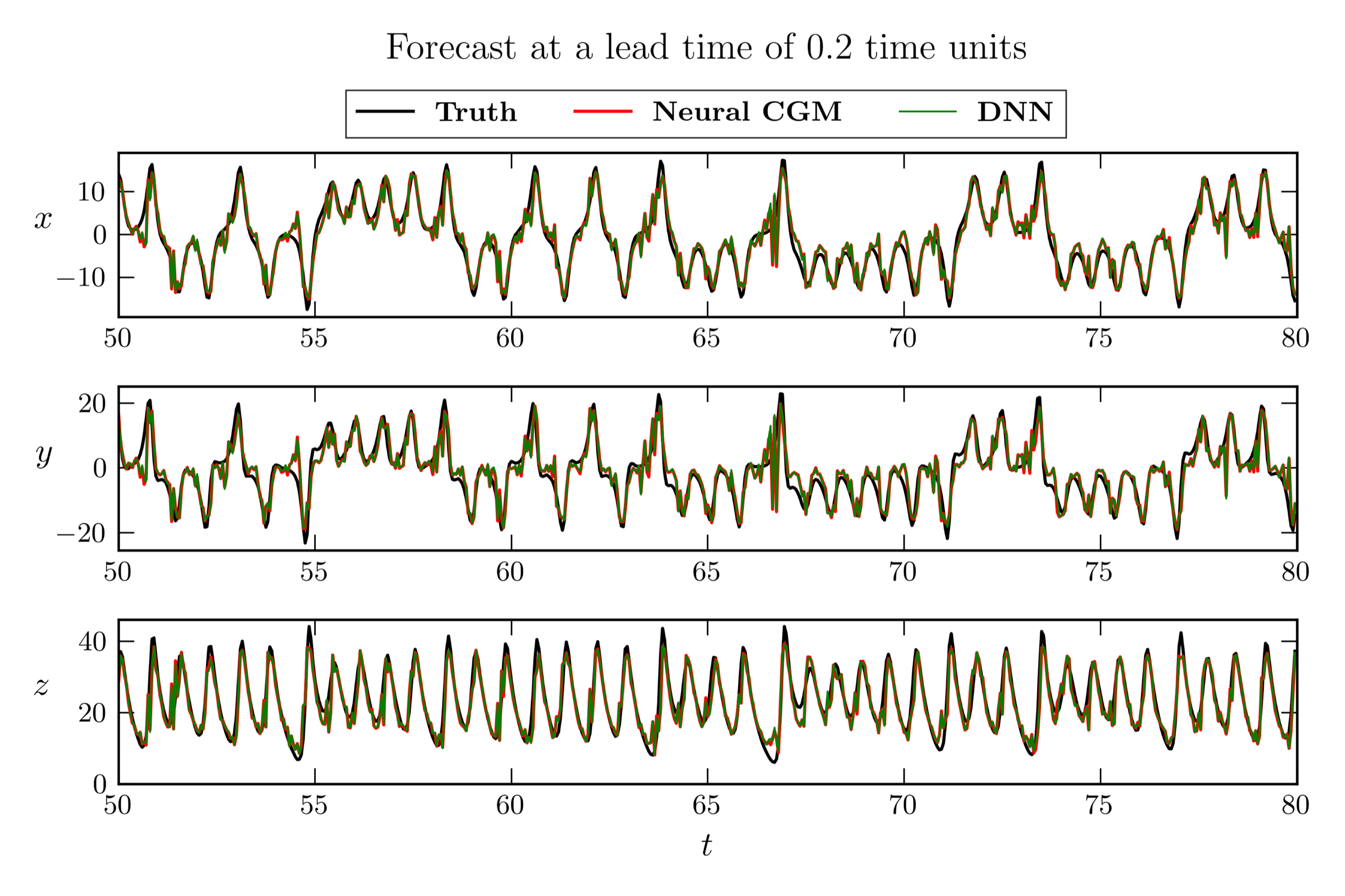}
    \caption{Time series of the state forecasting results for the stochastic Lorenz 63 system. The black trajectories denote the true states, while the red and green trajectories represent the predictions from the Neural CGM and DNN, respectively, at a lead time of $0.2$ time units.}
    \label{fig:L63_SF_TimeSeries}
\end{figure}

For state forecasting performance, the Neural CGM is compared with a deep neural network (DNN) of comparable model complexity. Figure~\ref{fig:L63_SF_TimeSeries} presents the state forecasting results of the Neural CGM and DNN at a lead time of $0.2$ time units. For this low-dimensional system, the DNN is sufficiently expressive to serve as a strong performance reference, while the Neural CGM achieves comparable forecasting accuracy. The close agreement with the DNN shows that Neural CGM retains enough expressive power to capture the underlying dynamics while also enabling efficient DA through the CG filter. In contrast, a general nonlinear DNN used as a dynamical model does not provide an analytical filtering update, so DA requires an additional nonlinear inference procedure, such as an ensemble-based method.

\subsection{Stochastic Lorenz 96 System}
\label{ssec:L96}

The Lorenz 96 system is a prototypical high-dimensional chaotic dynamical system and is widely used as a benchmark for nonlinear state estimation and data assimilation. In this example, we consider a stochastic 40-dimensional Lorenz 96 system:
\begin{equation}
\begin{aligned}
    \mathrm{d}u_i
    &=
    \left[
        (u_{i+1}-u_{i-2})u_{i-1}
        -u_i+F
    \right]\mathrm{d}t
    +
    \sigma\,\mathrm{d}W_i,
    \qquad i=1,\ldots,I,
\end{aligned}
\label{eq:L96}
\end{equation}
where \(u_i\) denotes the \(i\)-th state variable and \(W_i\) are independent standard Wiener processes. The system parameters are set as $I=40, F=8, \sigma=0.5$. The system is simulated over \(L_t=5000\) time units with a numerical integration time step of \(\Delta t=10^{-3}\) using the Euler-Maruyama method with initial condition $u(0)=0$. The simulated trajectory is sub-sampled every 50 numerical time steps, yielding observational data with temporal resolution $\Delta t_{\mathrm{obs}}=0.05$. Independent measurement noise following \(\mathcal{N}(0,0.75^2)\) is then added to all state variables. The first \(80\%\) of the trajectory is used as training data and the remaining \(20\%\) is used as test data. In the partially observed setting, four evenly spaced state variables, indexed by 1, 11, 21, and 31, are observed, while the remaining 36 state variables are treated as unobserved during DA. During DA evaluation, only the noisy observations of the four designated observed variables are provided, while the inferred unobserved states are compared against their corresponding noise-free ground truth.

In this case, the data-driven methods, Neural CGM and CGKN, and the classical DA methods, including EKF, UKF, EnKF, and PF, are benchmarked in terms of data assimilation accuracy and computational efficiency. The two data-driven methods are configured with comparable numbers of trainable model parameters to ensure a fair comparison. For the classical DA methods, the true governing equations of the Lorenz 96 system are assumed to be known and are integrated using an internal time step of $10^{-3}$ to ensure numerical stability and accuracy. For the UKF, the sigma-point parameters are set to $\alpha=0.72$, $\kappa=0$, and $\beta=2$. The EnKF employs $200$ ensemble members, with the standard deviation of observational noise inflated by a factor of $2$ relative to the original observation-noise standard deviation. The PF uses $1000$ particles, with resampling triggered when the effective sample size falls below $70\%$ of the particle population. A roughening coefficient of 0.01 controls the magnitude of Gaussian perturbations added to particles after resampling to mitigate particle impoverishment~\citep{gordon1993novel}. The standard deviation of the observation noise used in the likelihood function is inflated by a factor of 20 relative to the original observation-noise standard deviation. To investigate the effect of likelihood inflation on PF performance, we conduct a sensitivity analysis using inflation factors of 1, 5, 10, 20, and 40. The corresponding DA RMSE values are 5.0579, 4.7663, 3.9196, 3.4697, and 3.5841, respectively, while the DA NLL values are 19567.6055, 903.8100, 121.1100, 95.6056, and 96.9320, respectively. An inflation factor of 20 achieves the lowest DA RMSE and NLL among the tested candidates. It should be noted that this likelihood inflation is a pragmatic numerical adjustment to improve the performance of the PF with a finite number of particles (1000 in this example). With likelihood inflation, the resulting PF targets the Bayesian posterior under the inflated observation noise rather than that under the original observation model.

\begin{figure}[htbp]
    \centering
    \includegraphics[width=\linewidth]{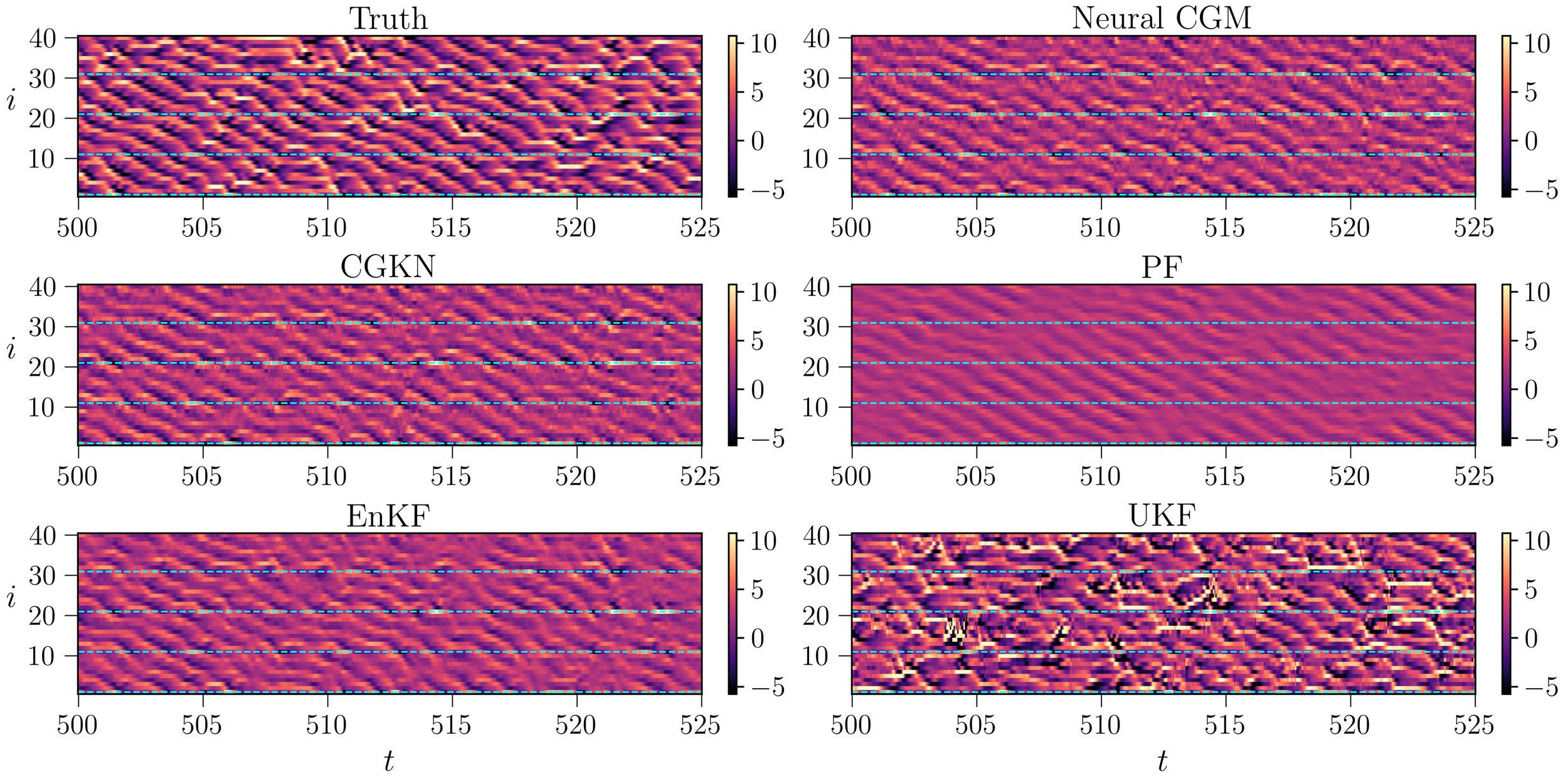}
    \caption{Data assimilation results for the Lorenz 96 system. The Truth panel shows the true state evolution, while the other panels display the posterior mean estimates produced by different DA methods. The blue dashed lines mark the indices of the observed state variables.}
    \label{fig:L96_DA_Pcolor}
\end{figure}

\begin{figure}[htbp]
    \centering
    \includegraphics[width=\linewidth]{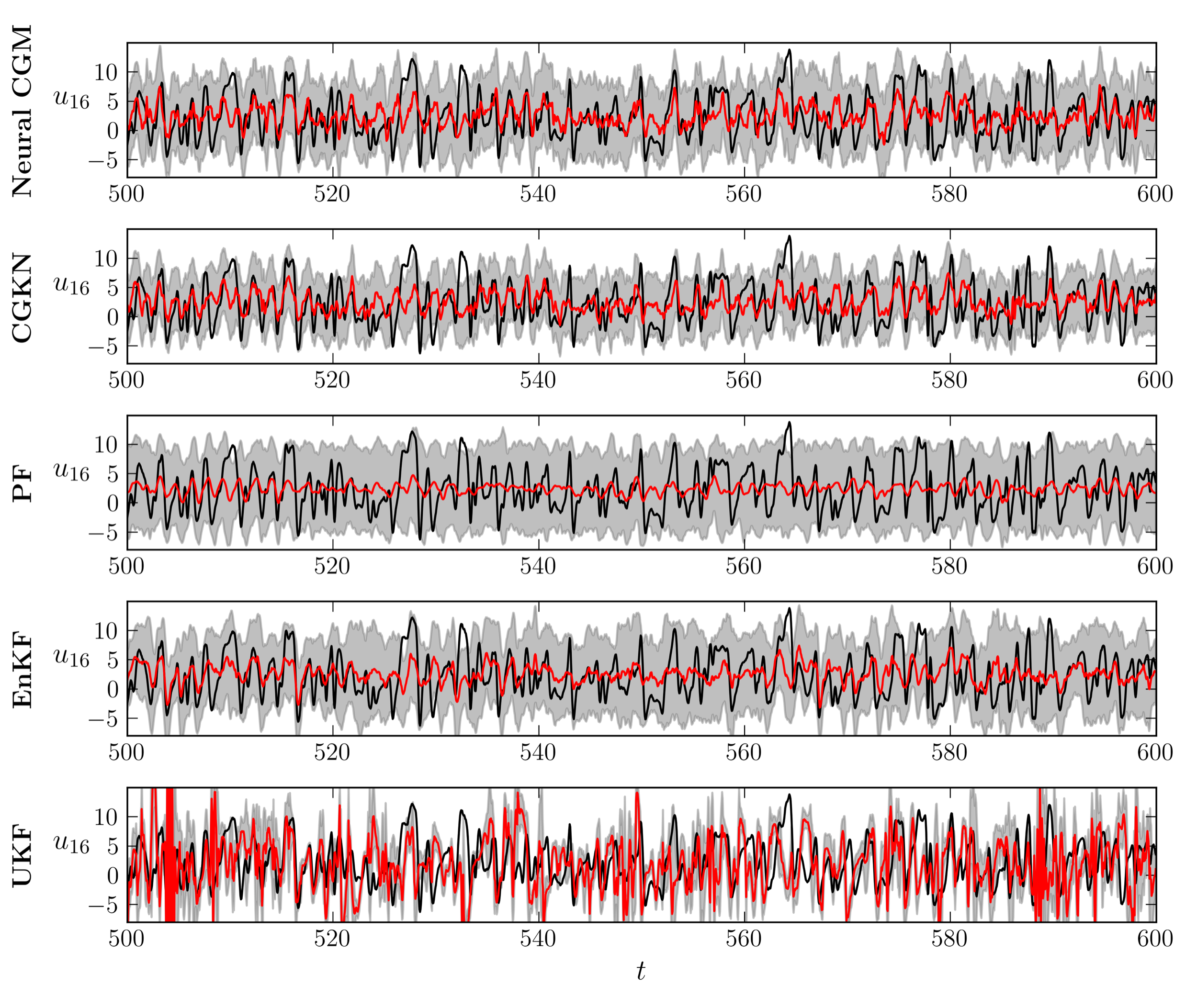}
    \caption{Time series of the data assimilation results for the stochastic Lorenz 96 system. The black trajectory represents the true signal of state variable $u_{16}$. The red trajectories represent the corresponding posterior means estimated by the various DA methods. The gray shaded regions indicate uncertainty intervals corresponding to $\pm 2$ posterior standard deviations.}
    \label{fig:L96_DA_TimeSeries}
\end{figure}

\begin{figure}[htbp]
    \centering
    \includegraphics[width=\linewidth]{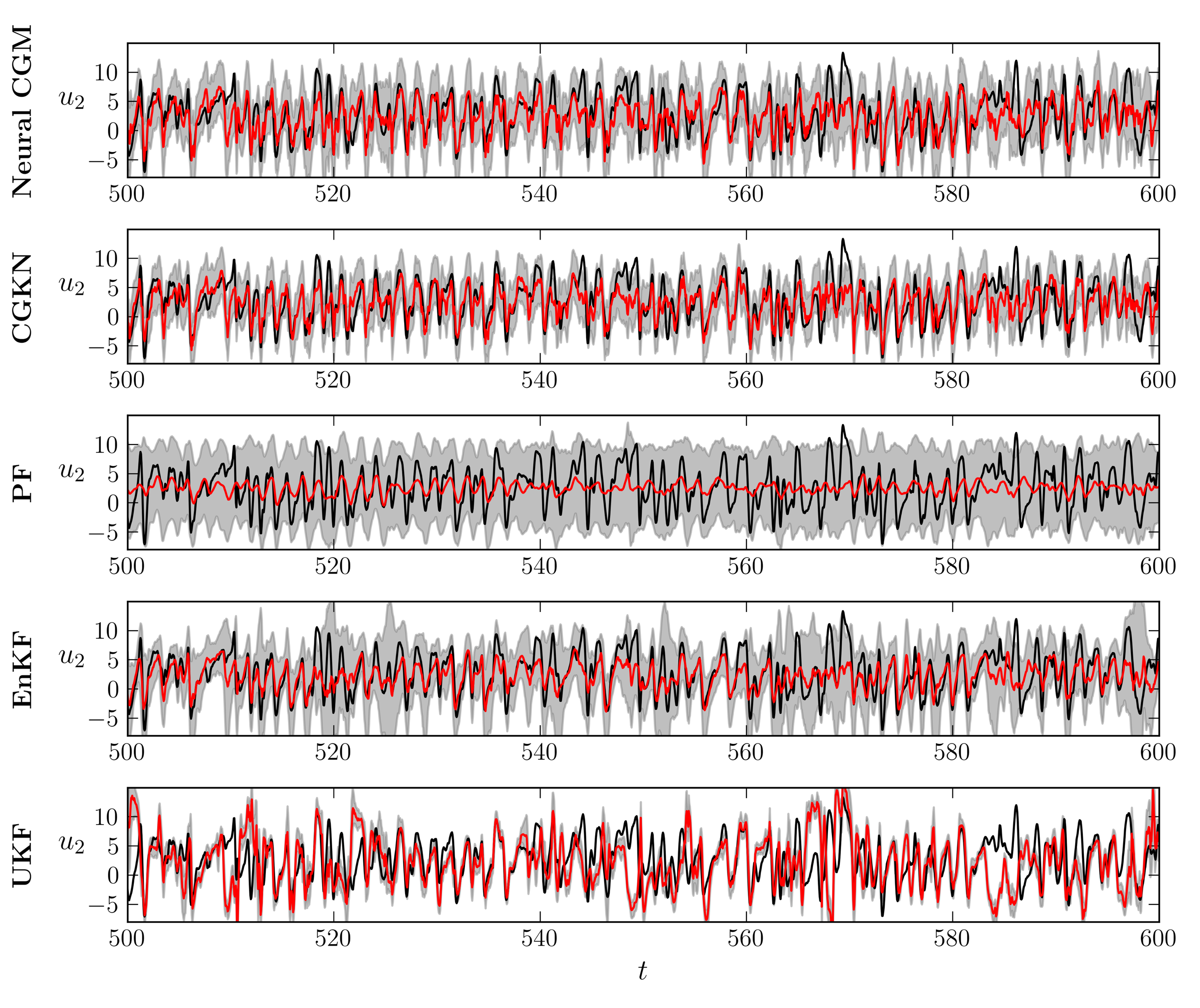}
    \caption{Time series of the data assimilation results for the stochastic Lorenz 96 system. The black trajectory represents the true signal of state variable $u_{2}$. The red trajectories represent the corresponding posterior means estimated by the various DA methods. The gray shaded regions indicate uncertainty intervals corresponding to $\pm 2$ posterior standard deviations.}
    \label{fig:L96_DA_TimeSeries_u2}
\end{figure}

Figures~\ref{fig:L96_DA_Pcolor}, \ref{fig:L96_DA_TimeSeries} and \ref{fig:L96_DA_TimeSeries_u2} illustrate the system evolution and the corresponding data assimilation results. Figure~\ref{fig:L96_DA_Pcolor} shows the spatiotemporal evolution of all state variables, where the four blue dashed trajectories indicate the indices of the observed states. The Truth panel represents the true system states, whereas the remaining panels show the posterior mean estimates obtained using different DA methods. Although the posterior means generally underestimate the amplitudes of the true states, Neural CGM, CGKN, and EnKF successfully capture the dominant spatiotemporal and chaotic patterns of the system. In contrast, the PF produces overly smooth estimates and fails to fully recover these patterns, while the UKF exhibits excessive fluctuations and less coherent state evolution. Figure~\ref{fig:L96_DA_TimeSeries} provides a more detailed comparison using the time series of the state variable $u_{16}$, corresponding to the state index $i=16$ in Figure~\ref{fig:L96_DA_Pcolor}. Neural CGM, CGKN, and EnKF show broadly similar behavior and are able to track the major temporal variations of the true state. The PF estimate is noticeably smoother and tends to suppress rapid fluctuations, whereas the UKF produces excessively oscillatory estimates. To complement the challenging case of $u_{16}$, which is relatively far from the observed variables, we also present the DA results for $u_2$ in Figure 3.5. Since $u_2$ is adjacent to an observed variable, it exhibits higher estimation accuracy than $u_{16}$. A quantitative summary of the DA performance of all methods is provided in Table~\ref{tab:benchmarks}.

\begin{figure}[htbp]
    \centering
    \includegraphics[width=0.9\linewidth]{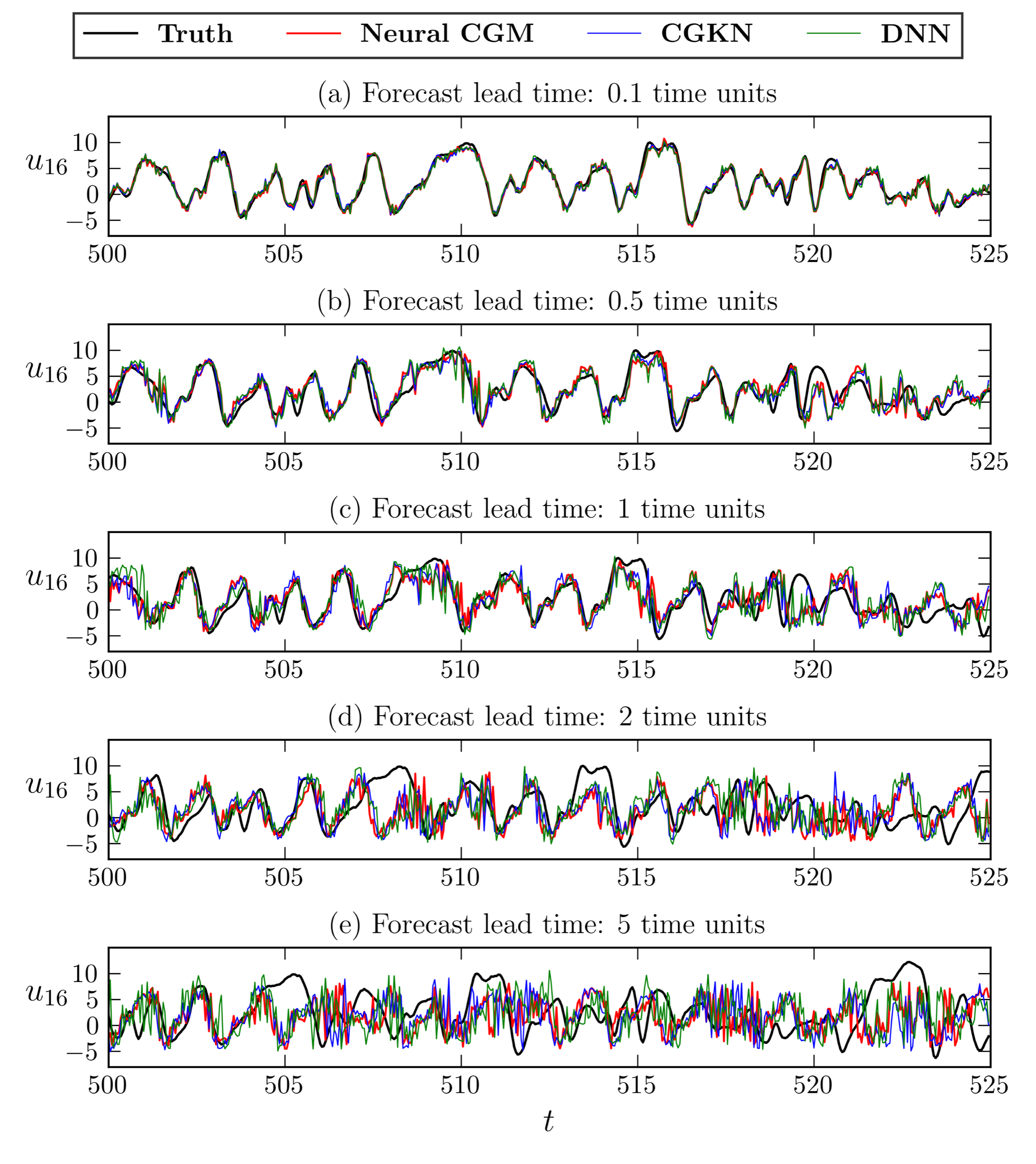}
    \caption{Time series of the state forecasting results for state variable of $u_{16}$ of the stochastic Lorenz 96 system. The black trajectory denotes the truth, while the red, blue and green trajectories represent the predictions from Neural CGM, CGKN, and DNN, respectively, at various lead times.}
    \label{fig:L96_SF_TimeSeries}
\end{figure}

For state forecasting, Neural CGM is compared with CGKN and DNN, all with similar numbers of model parameters. Figure~\ref{fig:L96_SF_TimeSeries} presents the forecasting results of the three models at lead times of \(0.1\), \(0.5\), \(1\), \(2\), and \(5\) time units. As the lead time increases, the forecasting accuracy of all three models gradually deteriorates due to the chaotic nature of the system. The DNN serves as a reference benchmark, demonstrating that both Neural CGM and CGKN achieve forecasting performance comparable to that of a general nonlinear neural-network model. The RMSEs of the one-step forecasts are reported in Table~\ref{tab:benchmarks}, where the three models exhibit very similar accuracy. This comparison shows that Neural CGM retains forecasting accuracy comparable to a general nonlinear neural model while also enabling efficient DA through the CG filter. In contrast, a general nonlinear DNN used as a dynamical model does not provide an analytical filtering update, so DA requires an additional nonlinear inference procedure to propagate uncertainty. Neural CGM yields slightly lower forecast RMSE, DA RMSE, and DA NLL than CGKN in this example, although the differences are small and should not be over-interpreted. In the next example, where the dynamics are higher-dimensional and spatiotemporal, the differences become much more pronounced.

\subsection{Kuramoto-Sivashinsky Equation}
\label{ssec:KSE}

The Kuramoto-Sivashinsky (KS) equation is a fourth-order nonlinear PDE used to model complex patterns and chaotic behavior in physical systems. It is commonly used as a benchmark system for studying chaos and data assimilation. The KS equation is given by
\begin{equation}
\begin{aligned}
    \frac{\partial u}{\partial t}
    =
    -u\frac{\partial u}{\partial x}
    -\frac{\partial^2 u}{\partial x^2}
    -\frac{\partial^4 u}{\partial x^4},
\end{aligned}
\label{eq:KSE}
\end{equation}
where $u(x, t)$ is the system state with $x \in (0,L_x)$ and $t \in (0,L_t]$, subject to periodic boundary conditions. The simulation settings are $L_x = 22, L_t = 50000, \Delta x = \frac{22}{2048}$, and $\Delta t = 0.025$. The initial condition is prescribed as $u(x,0) = 0.1\cos(x/16)\left(1+2\sin(x/16)\right)$. Observational noise is added independently to the simulation data with noise standard deviation $\sigma_{\mathrm{obs}} = 0.15\,\mathrm{std}(u)$. For model training and testing, the data are further sub-sampled to temporal resolution $\Delta t = 1$ and spatial resolution $\Delta x = 22/128$, resulting in 128 spatial states and 50000 time steps. The first $80\%$ of the trajectory is used as training data and the remaining $20\%$ is used as test data. Two spatial locations, indexed by $1$ and $65$, are observed, while the remaining 126 spatial states are treated as unobserved.

The Neural CGM is compared with CGKN and other classical DA methods, including EKF, UKF, EnKF, and PF, in terms of data assimilation accuracy and computational efficiency. Neural CGM and CGKN are configured with comparable numbers of trainable model parameters. For the classical DA methods, the true governing equations of the KS system are assumed to be known and are integrated using an internal time step of $0.025$ to ensure numerical stability and accuracy. For the EKF, the standard deviation of the observation noise is inflated by a factor of $15$ relative to the original observation-noise standard deviation. For the UKF, the sigma-point parameters are set to $\alpha=0.6$, $\kappa=0$, and $\beta=2$. The EnKF employs $500$ ensemble members, while the PF uses $2000$ particles, with resampling triggered when the effective sample size falls below $70\%$ of the number of particles. A roughening coefficient of $0.01$ is applied after resampling to mitigate particle impoverishment. With inflation factors of 1, 3, 5, 8, and 10 in PF, the corresponding DA RMSE values are 1.4819, 0.3966, 0.4347, 0.5596, and 0.6625, respectively, while the DA NLL values are 5959140.00, 117.0576, 52.4199, 86.7767, and 109.0935, respectively. The results for an inflation factor of 5 are reported in Table~\ref{tab:benchmarks}. In addition, since the KS system is a continuous spatiotemporal system governed by a PDE, spatial interpolation is included as a basic baseline for reference.

\begin{figure}[htbp]
    \centering
    \includegraphics[width=\linewidth]{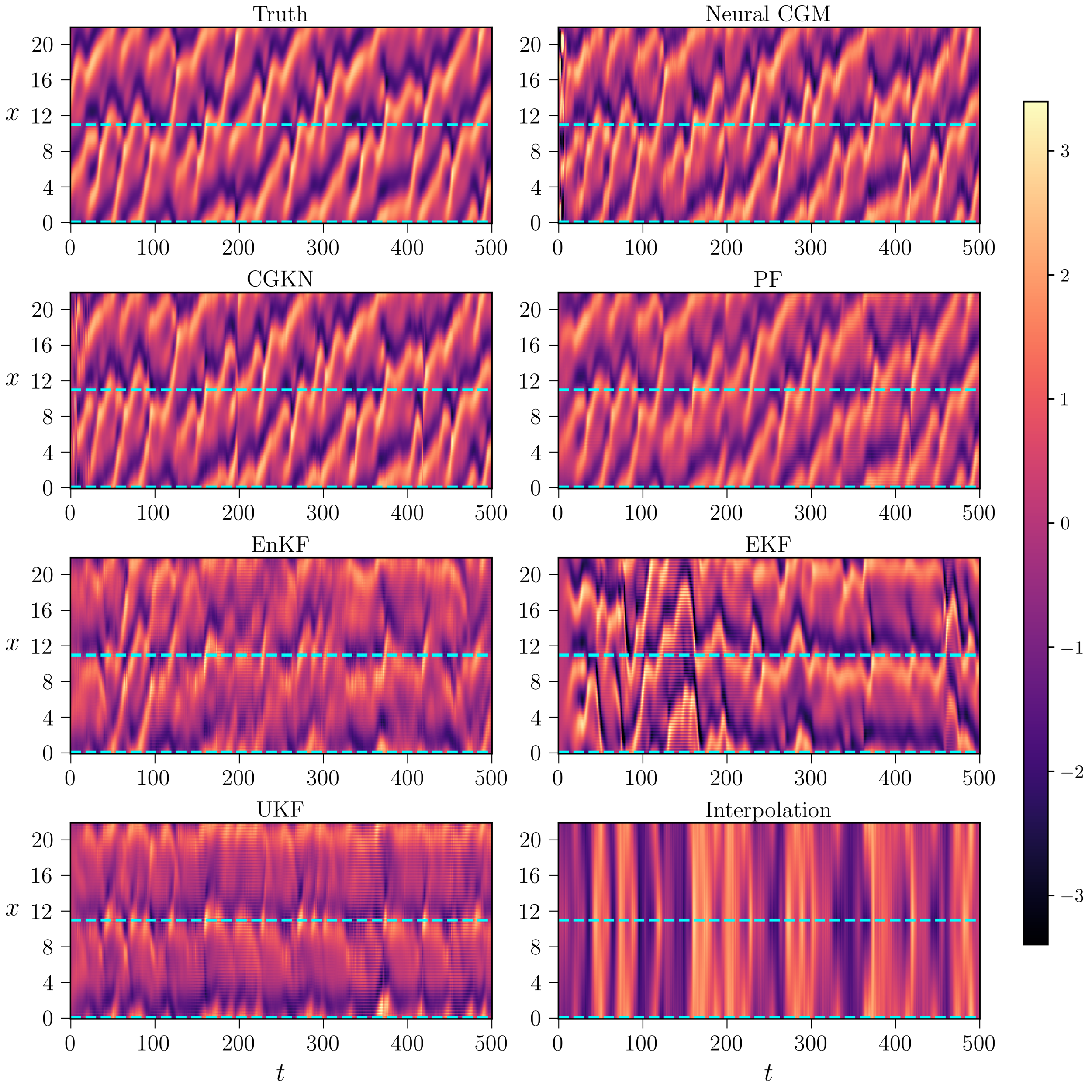}
    \caption{Data assimilation results for the Kuramoto-Sivashinsky equation. The Truth panel shows the true state evolution, while the other panels display the posterior mean estimates produced by different DA methods and the interpolation result. The blue dashed lines mark the indices of the observed spatial state.}
    \label{fig:KSE_DA_Pcolor}
\end{figure}

\begin{figure}[htbp]
    \centering
    \includegraphics[width=\linewidth]{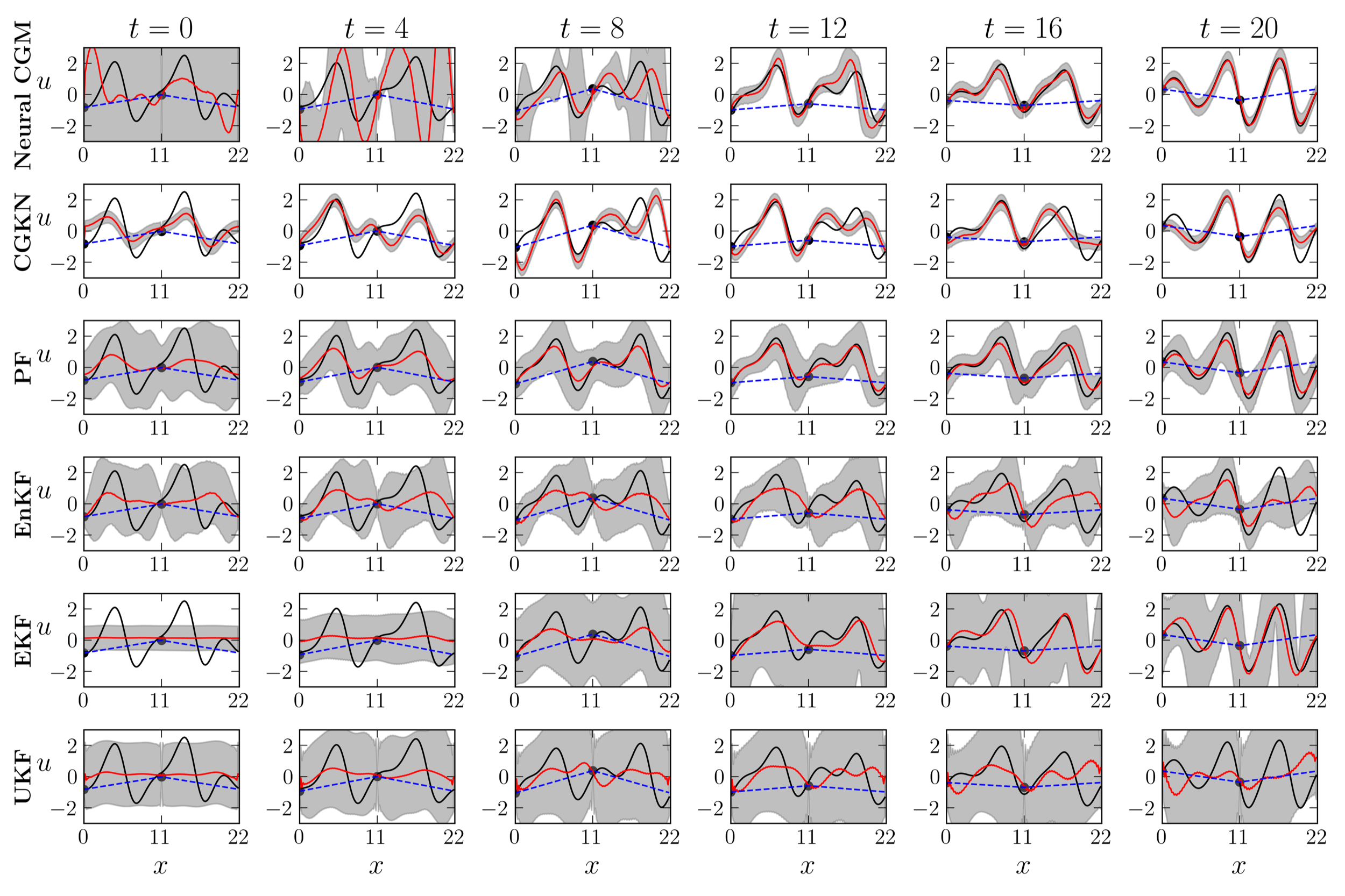}
    \caption{Data assimilation results for the Kuramoto–Sivashinsky equation from various methods. The black lines represent the true spatial states, while the red lines denote the posterior means estimated from each DA method. The gray shaded regions indicate the posterior mean \(\pm 2\) posterior standard deviations. The dashed blue lines show the spatial interpolation results based on the observations and are included as a baseline for reference.}
    \label{fig:KSE_DA_SpatialProfile}
\end{figure}

Figures~\ref{fig:KSE_DA_Pcolor} and \ref{fig:KSE_DA_SpatialProfile} present the data assimilation results from two complementary perspectives: the spatiotemporal evolution and the spatial profiles at selected time instances. In the Truth panel of Figure~\ref{fig:KSE_DA_Pcolor}, clear wave-like structures and chaotic dynamics can be observed, while the two dashed blue lines indicate the locations of the observed states. The remaining panels show the evolution of the posterior means obtained from different DA methods, together with the spatial interpolation baseline. Neural CGM, CGKN, and PF are able to preserve the overall wave-like and chaotic features of the true system, whereas EnKF, EKF, UKF, and spatial interpolation exhibit more substantial deviations from the truth.

Figure~\ref{fig:KSE_DA_SpatialProfile} provides a more detailed comparison through the corresponding spatial profiles at selected time instances. At the beginning of the assimilation process, all DA methods require some time to adjust from their initial estimates, corresponding to a DA warm-up period. By $t=12$, the posterior means produced by Neural CGM and PF closely capture the true spatial state, with the truth lying largely within their posterior uncertainty bands. CGKN also achieves comparable performance, whereas the other methods still exhibit noticeable discrepancies. By $t=20$, Neural CGM and PF provide the most accurate reconstructions of the true state among the compared methods, with uncertainty bands that cover much of the reference state. More specifically, it shows that Neural CGM produces narrower posterior uncertainty intervals than PF while achieving a lower DA RMSE, suggesting a more concentrated posterior estimate with improved state estimation accuracy. The Neural CGM requires only approximately $1.8$ s to complete the whole assimilation process for test data (10000 time steps), compared with $706.2211$ s for PF, making Neural CGM approximately $400$ times faster. It should be noted that the reported speedup depends on the numerical simulation and implementation of PF. In particular, the PF requires numerical integration of the KS dynamics for each particle between observations, whereas Neural CGM directly learns the state transition over the time interval between observations. More detailed quantitative comparisons are reported in Table~\ref{tab:benchmarks}. The dashed blue lines in Figure~\ref{fig:KSE_DA_SpatialProfile} represent the spatial interpolation results based solely on the observations and are included as a baseline for reference. The substantial improvement over spatial interpolation further demonstrates the importance of data assimilation for reconstructing the unobserved states.

\begin{figure}[htbp]
    \centering
    \includegraphics[width=\linewidth]{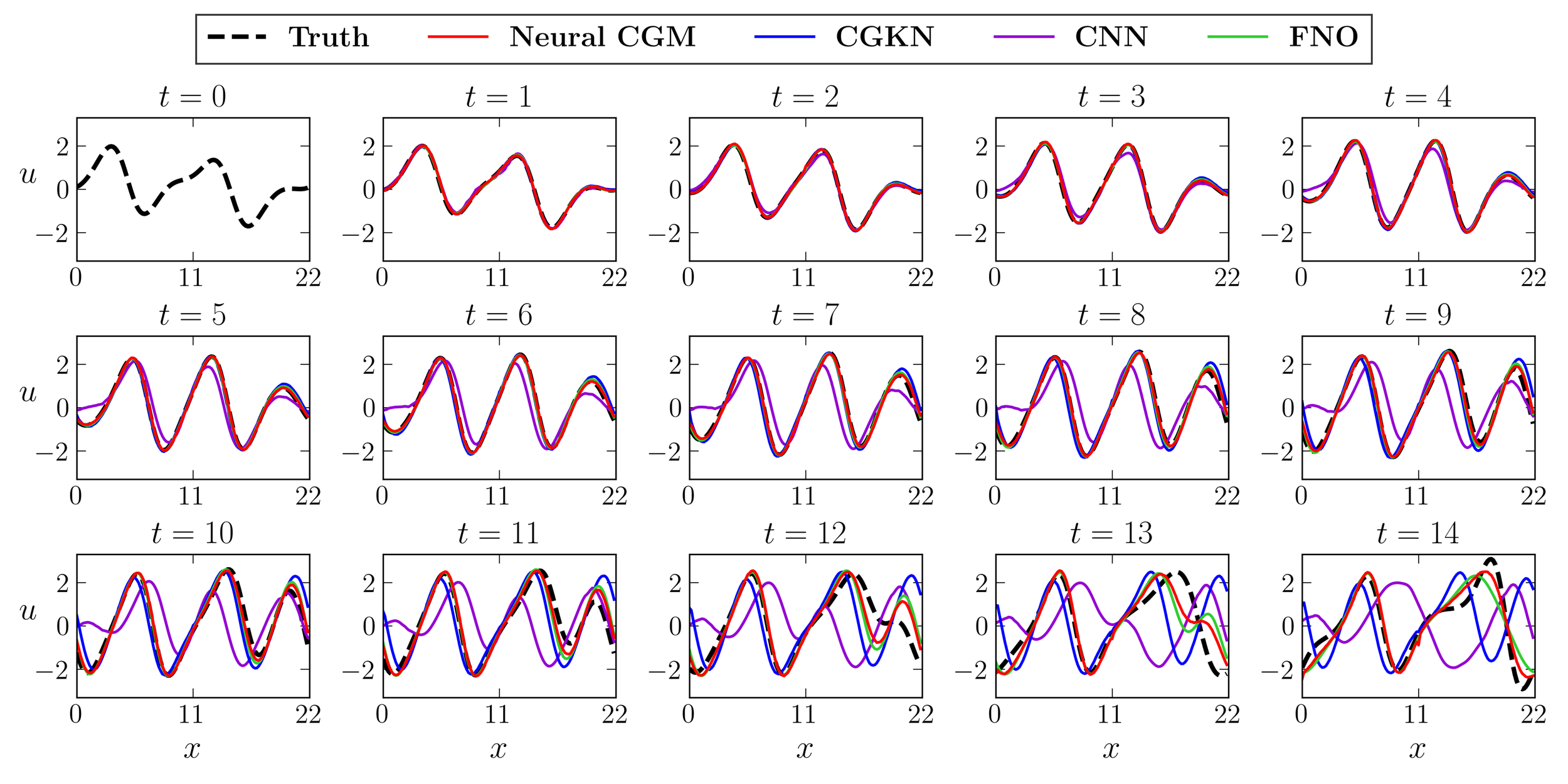}
    \caption{Results of state forecast for the Kuramoto-Sivashinsky equation. The evolution of true spatial functions is compared with predictive spatial functions from various models including Neural CGM, CGKN, CNN, and FNO.}
    \label{fig:KSE_SF_SpatialProfile}
\end{figure}

For state forecasting of this chaotic spatiotemporal dynamical system, the Neural CGM is compared with CGKN, CNN, and the Fourier Neural Operator (FNO), all configured with similar numbers of model parameters. Figure~\ref{fig:KSE_SF_SpatialProfile} presents the predicted spatial profiles over a forecast horizon of 14 time units, with the true spatial profile shown by the dashed black line. As the forecast time step increases, the predictions from all models gradually deviate from the truth due to the chaotic nature of the system. Among the compared models, CNN exhibits the earliest noticeable deviation, beginning at approximately $t=8$. The FNO serves as a reference benchmark, demonstrating that the Neural CGM can achieve forecasting performance comparable to that of an expressive neural architecture designed for solving nonlinear PDEs. The RMSEs of the one-step forecasts are reported in Table~\ref{tab:benchmarks}, where FNO and Neural CGM achieve the two lowest errors. This comparison shows that Neural CGM retains strong forecasting accuracy while also enabling efficient DA through the CG filter. In contrast, although FNO provides strong forecasting performance, it does not provide an analytical filtering update; using it as a dynamical model for DA therefore requires an additional nonlinear inference procedure to propagate uncertainty. Compared with CGKN, the Neural CGM exhibits better forecasting performance, with its predicted spatial profile remaining closer to the truth at $t=14$ and reaching a level of accuracy comparable to that of FNO.

\begin{figure}[htbp]
    \centering
    \includegraphics[width=0.8\linewidth]{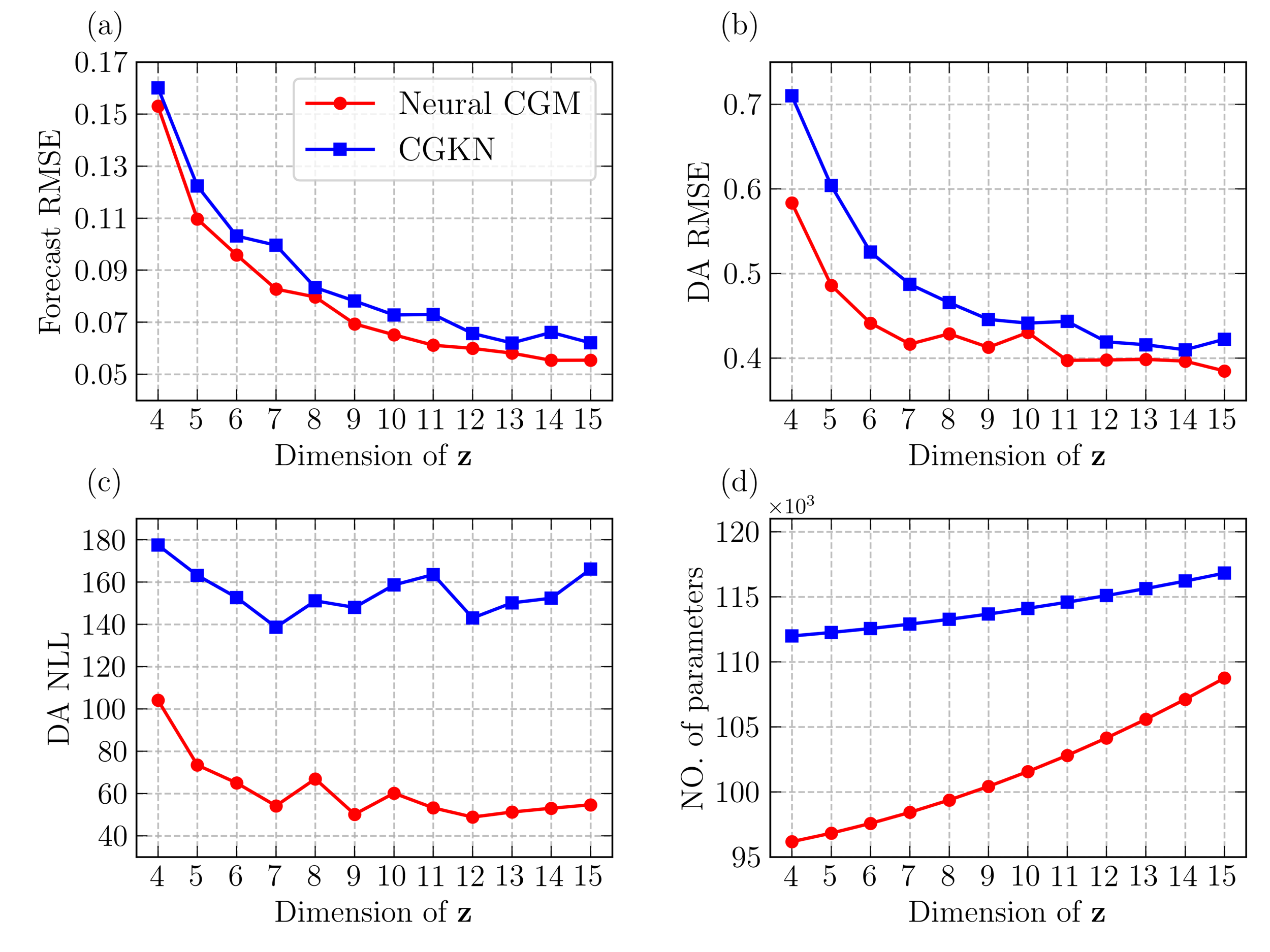}
    \caption{Comparison of Neural CGM and CGKN for the Kuramoto-Sivashinsky equation across different choices of the latent-state dimension $d_{\mathbf{z}}$. (a) Forecast RMSE, (b) DA RMSE, (c) DA negative log-likelihood (NLL), and (d) number of trainable parameters as functions of the latent-state dimension.}
    \label{fig:KSE_dim_z}
\end{figure}

Figure~\ref{fig:KSE_dim_z} compares Neural CGM and CGKN in terms of forecast RMSE, DA RMSE, DA negative log-likelihood (NLL), and the number of trainable parameters as functions of the latent-state dimension $\mathbf{z}$. In this case, Neural CGM yields lower errors and NLL than CGKN across the considered latent dimensions. The combined changes in decoder architecture, sequential covariance propagation, and likelihood-based DA training are associated with improved state forecasting, data assimilation, and uncertainty quantification while using fewer model parameters in these experiments. The numerical results reported in Table~\ref{tab:benchmarks} correspond to a latent state dimension of $d_{\mathbf{z}}=15$. These results show consistent gains over CGKN across the latent dimensions considered here while preserving a comparable model size. The improvement in DA negative log-likelihood is particularly pronounced for the KS system. Neural CGM reduces the DA RMSE relative to CGKN from \(0.4222\) to \(0.3846\), while the NLL drops much more sharply, from \(166.11\) to \(54.67\). Because the NLL depends on both the posterior mean and covariance, the larger change in NLL points to an improvement in the posterior uncertainty representation, not only in the mean estimate. The Neural CGM NLL is close to that of the particle filter, while the DA calculation requires only \(1.83\) s compared with \(706.22\) s for the particle filter. The same qualitative advantage is seen across the latent dimensions in Figure~\ref{fig:KSE_dim_z}, so the result is not tied to a single choice of latent dimension.

\section{Conclusions}
\label{sec:conclusion}

We developed the Neural Conditional Gaussian Model (Neural CGM), a data-driven model for nonlinear data assimilation (DA), state forecasting, and uncertainty quantification. Neural CGM learns conditional Gaussian dynamics in a latent space, where the posterior mean and covariance can be updated analytically, and maps these posterior statistics back to the physical variables through a distribution-aware Gaussian decoder. This construction separates point-state forecasting from posterior inference: the forecast decoder maps predictive latent states to physical states, whereas the DA decoder acts on both the latent posterior mean and covariance. The same posterior statistics enter the training objective through a Gaussian negative log-likelihood. Uncertainty quantification therefore remains part of the sequential DA procedure rather than being introduced only after state estimation.

The numerical experiments show that Neural CGM retains strong forecasting capability while substantially reducing the online cost of nonlinear DA. The benefit of explicitly modeling posterior uncertainty becomes particularly visible in the higher-dimensional Kuramoto-Sivashinsky example. Compared with CGKN, the improvement in posterior-mean RMSE is moderate, whereas the DA negative log-likelihood decreases from $166.11$ to $54.67$. Since the likelihood depends on both the posterior mean and covariance, this result suggests that the distribution-aware decoder and likelihood-based training improve the representation of the DA posterior beyond the state estimate alone. The resulting NLL is also close to that of the particle filter, while the corresponding online DA calculation is more than two orders of magnitude faster. Across the examples considered here, Neural CGM therefore combines accurate state estimation and forecasting with efficient sequential uncertainty propagation, without requiring explicit knowledge of the governing equations.

Several limitations remain. Training Neural CGM involves reconstruction, forecasting, and DA objectives whose relative weights may require tuning, and robust solutions can require multiple training runs or multi-stage strategies. The current formulation also assumes that paired trajectories of the variables treated as observed and unobserved are available during offline training, even though only the observed variables are available during online DA. Relaxing this requirement to partially supervised or heterogeneous training data is an important direction for future work. In addition, the present implementation uses only diagonal physical-space posterior covariances in the likelihood. Extending the framework to structured covariance representations would allow cross-variable posterior dependence to be incorporated without requiring a fully dense covariance model.

Neural CGM is most useful when state estimation, uncertainty quantification, and forecasting must be repeated from partial observations. Its latent representation can reduce the dimension on which the analytical DA is carried out, while the learned dynamical model avoids repeated integration of an unavailable, imperfect, or computationally expensive physical model. Larger and more complex systems are a natural next target, including numerical weather prediction, reconstruction of turbulent flows from sparse observations, and navigation problems with incomplete measurements.

\section*{Data and Code Availability}

The source code and data supporting the findings of this study are available at \url{https://github.com/Chuanqi-Chen/NeuralCGM}. The repository includes scripts for data generation, model training, evaluation, and figure reproduction.

\section*{Acknowledgement}
N.C. is supported by the Office of Naval Research under Award No. N00014-24-1-2244 and by the Army Research Office under Award No. W911NF-23-1-0118.

\bibliographystyle{unsrt}
\bibliography{references}


\end{document}